\documentclass{article}

\usepackage{microtype}
\usepackage{graphicx}
\usepackage{subfigure}
\usepackage{booktabs} 

\usepackage{xcolor}

\usepackage{amsmath,amsfonts,bm}

\def\eqref#1{equation~\ref{#1}}

\def\1{\bm{1}}

\DeclareMathAlphabet{\mathsfit}{\encodingdefault}{\sfdefault}{m}{sl}
\SetMathAlphabet{\mathsfit}{bold}{\encodingdefault}{\sfdefault}{bx}{n}

\usepackage{hyperref}
\usepackage{url}

\usepackage{booktabs}       
\usepackage{amsfonts}       
\usepackage{nicefrac}       
\usepackage{microtype}      

\usepackage[skip=3pt]{caption}

\usepackage{placeins}
\usepackage{amsmath}
\usepackage{amssymb}
\usepackage{graphicx}
\usepackage{mleftright}
\usepackage{enumitem}
\usepackage{float}

\usepackage{wrapfig}

\usepackage{booktabs,dcolumn}

\usepackage{siunitx}
\usepackage{tikz}
\usetikzlibrary{decorations.pathreplacing,calc}

\newcolumntype{d}[1]{D{.}{.}{#1}}

\newcommand*\mcTwo[1]{\multicolumn{2}{l}{#1}}

\usepackage{hyperref}

\usepackage[noend, ruled,vlined]{algorithm2e}

\usepackage[final]{corl_2021} 

\title{Learning to Plan Optimistically: Uncertainty-Guided Deep Exploration via Latent Model Ensembles}

\author{
Tim Seyde\thanks{Equal contribution. Correspondence to \texttt{\{tseyde,wilkos\}@mit.edu}}\\
MIT CSAIL \\
\And
Wilko Schwarting$^*$\\
MIT CSAIL \\
\And
Sertac Karaman\\
MIT LIDS \\
\And
Daniela Rus\\
MIT CSAIL
}

\begin{document}
\maketitle


\begin{abstract}
    Learning complex robot behaviors through interaction requires structured exploration. 
    Planning should target interactions with the potential to optimize long-term performance, while only reducing uncertainty where conducive to this objective.
    This paper presents Latent Optimistic Value Exploration (LOVE), a strategy that enables deep exploration through optimism in the face of uncertain long-term rewards.
    We combine latent world models with value function estimation to predict infinite-horizon returns and recover associated uncertainty via ensembling.
    The policy is then trained on an upper confidence bound (UCB) objective to identify and select the interactions most promising to improve long-term performance.
    We apply LOVE to visual robot control tasks in continuous action spaces and demonstrate on average more than 20\% improved sample efficiency in comparison to state-of-the-art and other exploration objectives. In sparse and hard to explore environments we achieve an average improvement of over 30\%.
\end{abstract}



\section{Introduction}
\begin{wrapfigure}{r}{0.50\textwidth}
\vspace{-35pt}
\begin{center}
  \includegraphics[width=1.00\linewidth]{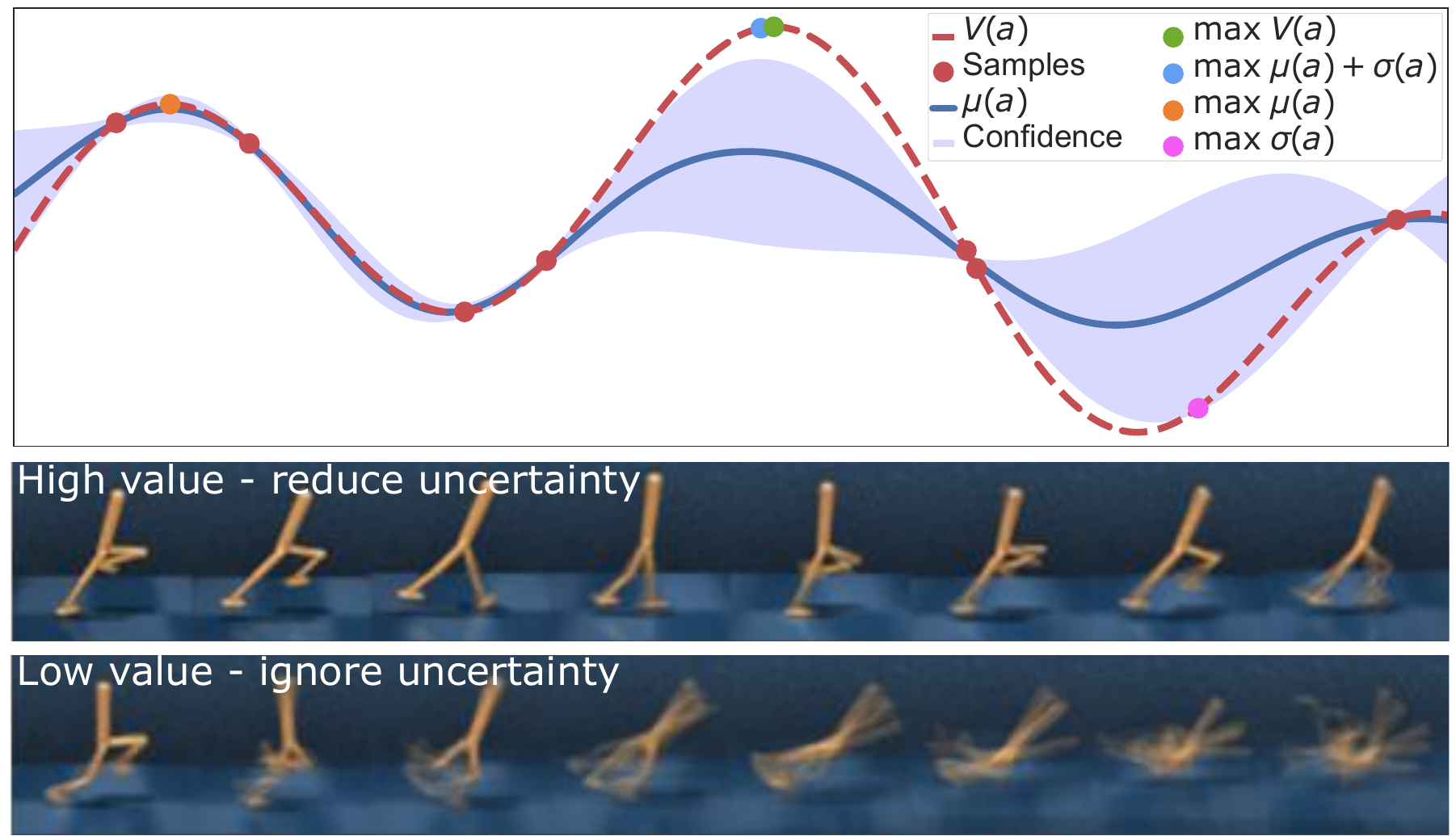}
\end{center}
\caption{
Top - Uncertainty over returns can guide exploration to regions with high potential for improvement (blue dot) during sampling-based value learning (red line). Bottom - uncertainty reduction should focus on expected high-reward behaviors.
}
\vspace{-20pt}
\label{fig:gaussian_process_regression}
\end{wrapfigure}%
%
The ability to learn complex behaviors through interaction will enable the autonomous deployment of various robotic systems in the real world.
Reinforcement learning (RL) provides a key framework for realizing these capabilities, but efficiency of the learning process remains a prevalent concern. 
Real-life applications yield complex planning problems due to high-dimensional environment states, which are further exacerbated by the agent's continuous action space.
For RL to enable real-world autonomy, it therefore becomes crucial to determine efficient representations of the underlying planning problem, while formulating interaction strategies capable of exploring these representations efficiently.
%
 
In traditional controls, planning problems are commonly formulated based on the underlying state-space representation. 
This may inhibit efficient learning when the environment states are high-dimensional or their dynamics are susceptible to non-smooth events such as singularities and discontinuities \citep{schrittwieser2019mastering, hwangbo2019learning, yang2019data}.
It may then be desirable for the agent to abstract a latent representation that facilitates efficient learning \citep{ha2018world, zhang2019solar, lee2019slac}.
The latent representation may then be leveraged either in a model-free or model-based setting.
Model-free techniques estimate state-values directly from observed data to distill a policy mapping. 
Model-based techniques learn an explicit representation of the environment that is leveraged in generating fictitious interactions and enable policy learning in imagination \citep{hafner2019dream}.
While the former reduces potential sources of bias, the latter offers a structured representation encoding deeper insights into underlying environment behavior.
%

The agent should leverage the chosen representation to efficiently identify and explore informative interactions.
For example, an engineer would ideally only provide sparse high-level feedback to a robot and the system should learn without getting distracted by environment behavior irrelevant to task completion. 
This requires models that can disentangle behavioral surprise from expected performance and motivate the agent to investigate the latter.
Figure~\ref{fig:gaussian_process_regression} provides a one-dimensional example of a value mapping.
The true function and its samples are visualized in red with the true maximum denoted by the green dot.
Relying only on the predicted mean can bias policy learning towards local optima (orange dot; \citet{sutton2018reinforcement}), while added stochasticity can waste samples on uninformative interactions.
Auxiliary information-gain objectives integrate predicted uncertainty, however, uncertain environment behavior does not equate to potential for improvement (pink dot).
It is desirable to focus exploration on interactions that harbor potential for improving overall performance. 
Combining mean performance estimates with their uncertainty into an upper confidence bound (UCB) objective provides a concise method for this (blue dot; \citet{auer2002ucb, krause2011ucb}).
The uncertainty can be explicitly represented by an ensemble of hypothesis on environment behavior \citep{osband2016deep, lakshminarayanan2017ensemble}.
Figure~\ref{fig:gaussian_process_regression} demonstrates this selective uncertainty reduction by providing forward predictions of a model ensemble on two motion patterns for a Walker. The expected high-reward walking behavior has been sufficiently explored and the models strongly agree, while little effort has been extended to reduce uncertainty and learn the specific details of the expected low-reward falling behavior.
%

This paper demonstrates that exploring interactions through imagined positive futures can yield information-dense sampling and data-efficient learning. We present latent optimistic value exploration (LOVE), an algorithm that leverages optimism in the face of uncertain long-term rewards in guiding exploration. Potential futures are imagined by an ensemble of latent variable models and their predicted infinite-horizon performance is obtained in combination with associated value function estimates. Training on a UCB objective over imagined futures yields a policy that behaves inherently optimistic and focuses the robot learner on interactions with the potential to improve performance under its current world model. This provides a concise, differentiable framework for driving deep exploration while not relying on stochasticity. LOVE therefore provides an agent with
\begin{itemize}[leftmargin=+1.5em]
    \item an uncertainty-aware model for diverse hypotheses on trajectory evolution and performance,
    \item intrinsic exploration without reward feedback by active querying of uncertain long-term returns,
    \item targeted exploration guided by improvement potential, ignoring uncertainty tangential to the task.
\end{itemize}
We present empirical results on challenging visual robot control tasks that highlight the necessity for deep exploration in scenarios with sparse rewards, where we achieve on average more than 30\% improvement over state-of-the-art and an exploration baseline. We further demonstrate $15\%$ average improvement on benchmarking environments from the DeepMind Control Suite \citep{deepmindcontrolsuite2018}. We compare to the recent agents Dreamer \citep{hafner2019dream}, a curiosity-based agent inspired by \citet{sekar2020planning}, and DrQ \citep{kostrikov2020image}.%

\section{Related work}
\paragraph{Problem representation}
Model-free approaches learn a policy by directly estimating performance from interaction data. 
While their asymptotic performance previously came at the cost of sample complexity \citep{lillicrap2015continuous, fujimoto2018td3, haarnoja2018sac}, recent advances in representation learning through contrastive methods and data augmentation have improved their efficiency \citep{srinivas2020curl, laskin2020reinforcement, kostrikov2020image}. 
However, their implicit representation of the world can make generalization of learned behaviors under changing task specifications difficult.
Model-based techniques leverage a structured representation of their environment that enables them to imagine potential interactions. 
The nature of the problem hereby dictates model complexity, ranging from linear \citep{levine2014learning, kumar2016optimal}, over Gaussian process models \citep{deisenroth2011pilco, kamthe2017data} to deep neural networks \citep{chua2018pets, clavera2018mbmpo}.
In high-dimensional environments, latent variable models can provide concise representations that improve efficiency of the learning process \citep{watter2015embed, ha2018world, lee2019slac, hafner2019dream}.

\paragraph{Planning interactions}
Model-based approaches leverage their representation of the world in predicting the performance of action sequences. The agent may then either solve a model predictive control-style optimization \citep{nagabandi2018mbmf, chua2018pets, hafner2018planet} or train a policy in simulation \citep{kurutach2018metrpo, clavera2018mbmpo}. 
The resulting finite-horizon formulations can be extended by value function estimates to approximate an infinite-horizon planning problem \citep{lowrey2018polo, hafner2019dream, seyde20dove}. 
When considering learned models, ensembling the model predictions may further be leveraged in debiasing the actor during training \citep{kurutach2018metrpo, chua2018pets, clavera2018mbmpo, seyde20dove}.
Both explicit and implicit model rollouts together with value estimation can accelerate model-free learning \citep{oh2017value, feinberg2018model, buckman2018sample}.

\paragraph{Directed exploration}
Directed exploration can improve over random exploration by focusing on information-dense interactions \citep{schmidhuber2010formal}. 
These methods are commonly driven by uncertainty estimates. Information gain techniques define an auxiliary objective that encourages exploration of unexpected environment behavior or model disagreement and have been applied in discrete \citep{stadie2015incentivizing, ostrovski2017count, pathak2017curiosity} and continuous actions spaces \citep{still2012information, houthooft2016vime, henaff2019explicit}.
Improving knowledge of the dynamics yields general purpose models, while the agent may explore uncertain interactions tangential to a specific task objective.
Alternatively, exploration can be driven by uncertainty over performance as encoded by value functions \citep{osband2016deep, osband2017deep, chen2017ucb, odonoghue2017uncertainty, lee2020sunrise}, multi-step imagined returns \citep{depeweg2018decomposition, henaff2019model} or their combination \citep{lowrey2018polo, schrittwieser2019mastering, seyde20dove}.

\paragraph{Model-ensemble agents}
Related work on ensemble agents has demonstrated impressive results. We note key differences to our approach.
ME-TRPO~\citep{kurutach2018metrpo} leverages a dynamics ensemble to debias policy optimization on finite-horizon returns under a known reward function and random exploration.
MAX~\citep{shyam2019model} and \citet{amos2018learning} explore via finite-horizon uncertainty in a dynamics ensemble. 
Plan2Explore~\citep{sekar2020planning} combines this idea with Dreamer~\citep{hafner2019dream}, learning a model that enables adaptation to multiple downstream tasks.
RP1~\citep{ball2020ready} explores in reward space via finite-horizon returns, but assumes access to the nominal reward function and full proprioceptive feedback.
\citet{seyde20dove} leverage full proprioception and embed optimism into the value function, which prohibits adjustment of the exploration trade-off during learning and limits transferability.
Concurrent work by~\citet{rafailov2021offline} penalizes uncertainty over latent states for learning robust visual control policies via offline RL.
Exploring uncertain dynamics samples interactions orthogonal to task completion and finite-horizon objectives limit exploration locally, while full-observability and access to the reward function are strong assumptions.
We learn latent dynamics, reward and value functions under partial observability to explore uncertain infinite-horizon returns. This enables backpropagation through imagined trajectories to recover analytic policy gradients, as well as guided deep exploration based on expected potential for long-term improvement.

\section{Preliminaries}
\label{sec:preliminaries}
\subsection{Problem formulation}
We formulate the underlying optimization problem as a partially observable Markov decision process (POMDP) defined by the tuple $\{\mathcal{X}, \mathcal{A}, T, R, \Omega, \mathcal{O}, \gamma\}$, where $\mathcal{X}$, $\mathcal{A}$, $\mathcal{O}$ denote the state, action and observation space, respectively, $T \colon \mathcal{X} \times \mathcal{A} \to \mathcal{X}$ signifies the transition mapping, $R \colon \mathcal{X} \times \mathcal{A} \to \mathbb{R}$ the reward mapping, $\Omega \colon \mathcal{X} \to \mathcal{O}$ the observation mapping, and $\gamma \in[0,1)$ is the discount factor. 
We define $x_t$ and $a_t$ to be the state and action at time $t$, respectively, and use the notation $r_t = R\mleft(x_t, a_t\mright)$. 
Let $\pi_{\phi} \mleft(a_t | o_t\mright)$ denote a policy parameterized by $\phi$ and define the discounted infinite horizon return $G_t = \sum_{\tau=t}^{\infty}\gamma^{\tau-t} R\mleft(x_{\tau}, a_{\tau}\mright)$, where $x_{t+1} \sim T\mleft(x_{t+1} | x_t, a_t\mright)$ and $a_t \sim \pi_{\phi}\mleft(a_t | o_t\mright)$. 
The goal is then to learn the optimal policy maximizing $G_t$ under unknown nominal dynamics and reward mappings.

\subsection{Planning from pixels}
\label{sec:dreamer}
We build on the world model introduced in \citet{hafner2018planet} and refined in \citet{hafner2019dream}. High-dimensional image observations are first embedded into a low-dimensional latent space using a neural network encoder. A recurrent state space model (RSSM) then provides probabilistic transitions and defines the model state $s$.
Together, the encoder and RSSM define the representation model. 
The agent therefore abstracts observation $o_t$ of environment state $x_t$ into model state $s_t$, which is leveraged for planning.
Consistency of the learned representations is enforced by minimizing the reconstruction error of a decoder network in the observation model and the ability to predict rewards of the reward model. For details, we refer the reader to \citet{hafner2019dream}, and provide their definitions
\begin{equation}
    \begin{aligned}
        & \text{Representation model:} && \qquad \qquad \textstyle p_{\theta} \mleft(s_{t} | s_{t-1}, a_{t-1}, o_t\mright) \\
        & \text{Transition model:} && \qquad \qquad \textstyle q_{\theta} \mleft(s_{t} | s_{t-1}, a_{t-1}\mright) \\
        & \text{Observation model:} && \qquad \qquad \textstyle q_{\theta} \mleft(o_{t} | s_{t}\mright) \\
        & \text{Reward model:} && \qquad \qquad \textstyle q_{\theta} \mleft(r_{t} | s_{t}\mright),
    \end{aligned}
\end{equation}
where $p$ and $q$ denote distributions in latent space, with $\theta$ as their joint parameterization.
The action model $\pi_{\phi}\mleft(a_{t} | s_{t}\mright)$ is then trained to optimize the predicted return of imagined world model rollouts. The world model is only rolled-out over a finite horizon $H$, but complemented by predictions from the value model $v_{\psi}\mleft(s_{t}\mright)$ at the terminal state $s_{t+H}$ to approximate the infinite horizon return. The policy and value function are trained jointly using policy iteration on the objective functions
\begin{equation}
    \begin{aligned}
        \textstyle & \max_{\phi} E_{q_{\theta}, q_{\phi}}\mleft(\sum_{\tau=t}^{t+H} V_{\lambda}\mleft(s_{\tau}\mright)\mright), \qquad
        \textstyle & \min_{\psi} E_{q_{\theta}, q_{\phi}}\mleft(\sum_{\tau=t}^{t+H} \frac{1}{2} \lVert v_{\psi}\mleft(s_{\tau}\mright) - V_{\lambda} \mleft(s_{\tau}\mright)\rVert ^{2}\mright),
    \end{aligned}
    \label{eq:dreamer_objectives}
\end{equation}
respectively. Here, $V_{\lambda}\mleft(s_{\tau}\mright)$ represents an exponentially recency-weighted average of the $k$-step value estimates $V_{N}^{k}\mleft(s_{\tau}\mright)$ along the trajectory to stabilize the learning \citep{sutton2018reinforcement} as further specified in Appendix~\ref{sec:lambda_return}.

\section{Uncertainty-guided latent exploration}
The world model introduced in Section \ref{sec:dreamer} can be leveraged in generating fictitious interactions for the policy to train on. However, the learned model will exhibit bias in uncertain regions where insufficient samples are available. Training on imagined model rollouts then propagates simulation bias into the policy. 
Here, we address model bias by ensembling our belief over environment behavior. We can leverage the underlying epistemic uncertainty in formulating a UCB objective for policy learning that focuses exploration on regions with high predicted potential for improvement.%
\subsection{Model learning with uncertainty estimation}
The model parameters are only weakly constrained in regions where interaction data is scarce and random influences strongly impact prediction performance.
In order to prevent the agent from learning to exploit these model mismatches, we consider predictions from an ensemble. 
Individual predictions will align in regions of high data support and diverge in regions of low support. 
The ensemble mean then serves as a debiased estimator of environment behavior, while the epistemic uncertainty is approximated via model disagreement \citep{lakshminarayanan2017ensemble}.
We consider an ensemble of $M$ latent-space particles. Each particle is represented by a unique pairing of a transition, reward and value model to yield,
\begin{equation}
    \begin{aligned}
        \text{Particles:} \,\, \textstyle \{\mleft(q_{\theta_{i}} \mleft(s_{t} | s_{t-1}, a_{t-1}\mright), \; q_{\theta_{i}} \mleft(r_{t} | s_{t}\mright), \; v_{\psi_i}\mleft(s_{t}\mright)\mright)\}_{i=1}^{M}.
    \end{aligned}
\end{equation}
The encoder and decoder remain shared to ensure that all particles leverage the same latent space, while the internal transition dynamics retain the ability of expressing distinct hypothesis over environment behavior. For particle $i$, we define the predicted infinite-horizon trajectory return as
\begin{align}
    G_{t,i}\mleft(\theta_i, \psi_i, \phi\mright) = \sum_{\tau=t}^{t+H} V_{\lambda, i}\mleft(s_{\tau}\mright),
\label{eq:particle_return}
\end{align}
where $V_{\lambda, i}\mleft(s_{\tau}\mright)$ is computed via Eq.~(\ref{eq:value_estimates}) with the particle's individual transition, reward and value models. Distinctness of the particles is encouraged by varying the initial network weights between ensemble members and shuffling the batch order during training.
Predicted trajectory returns with corresponding uncertainty estimates are then obtained by considering the ensemble mean and variance,
\begin{align}
    \mu_{G}\mleft(\theta, \psi, \phi\mright) &= \frac{1}{M} \sum_{i=1}^{M} G_{t,i}\mleft(\theta_i, \psi_i, \phi\mright), \label{eq:mean_return}\\
    \sigma_{G}^{2}\mleft(\theta, \psi, \phi\mright) &= \frac{1}{M}\sum_{i=1}^{M}\mleft(G_{t,i}\mleft(\theta_i, \psi_i, \phi\mright) - \mu_{G}\mleft(\theta, \psi, \phi\mright)\mright)^{2}.
\label{eq:rollout_moments}
\end{align}%
\subsection{Policy learning with directed exploration}
The policy learning objective in Eq.~(\ref{eq:dreamer_objectives}) could be replaced by the ensemble mean in Eq.~(\ref{eq:rollout_moments}). This would reduce model bias in the policy, but require an auxiliary objective to ensure sufficient exploration. We consider exploration to be desirable when it reduces uncertainty over realizable task performance. The trajectory return variance in Eq.~(\ref{eq:rollout_moments}) encodes uncertainty over long-term performance of actions. In combination with the expected mean, we recover estimated performance bounds. During data acquisition, we explicitly leverage the epistemic uncertainty in identifying interactions with potential for improvement and define the acquisition policy objective via an upper confidence bound (UCB) as%
\begin{equation}
\label{eq:ucb_objective_aquisition}
    \begin{aligned}
    G_{aq}\mleft(\theta, \psi, \phi\mright) = \mu_{G}\mleft(\theta, \psi, \phi\mright) + \beta \sigma_{G}\mleft(\theta, \psi, \phi\mright),
\end{aligned}
\end{equation}%
where the scalar variable $\beta$ quantifies the exploration-exploitation trade-off. For $\beta < 0$ we recover a safe-interaction objective, while $\beta > 0$ translates to an inherent optimism that uncertainty harbors potential for improvement. Here, we learn an optimistic policy $\pi_{\phi_{aq}}$ that is intrinsically capable of deep exploration and focuses interactions on regions with high information-density. Furthermore, in the absence of dense reward signals, the acquisition policy can leverage prediction uncertainty in driving exploration. This behavior is not limited to the preview window, as the value function ensemble projects long-term uncertainty into the finite-horizon model rollouts. While training in imagination, we leverage the optimistic policy to update our belief in regions that the acquisition policy intends to visit. In parallel, we train an evaluation policy $\pi_{\phi_{ev}}$ that aims to select the optimal actions under our current belief. The evaluation policy optimizes the expected mean return ($\beta=0$).%
\subsection{Latent optimistic value exploration (LOVE)}
In the following, we provide a high-level description of the proposed algorithm, LOVE, together with implementation details of the overall training process and the associated pseudo-code in Algorithm~\ref{alg:love}.
\paragraph{Summary}
\begin{wrapfigure}[21]{r}{0.53\textwidth} 
    \vspace{-15pt}
  \begin{algorithm}[H]                
    \small
    \SetCustomAlgoRuledWidth{0.47\textwidth}  
      \SetInd{0.3em}{0.3em}
    \SetAlgoLined\DontPrintSemicolon
    \SetKwInOut{Initialize}{Initialize}
    \SetKwFunction{rollout}{rollout}
    \SetKwProg{myalg}{Algorithm}{}{}
        \Initialize{random parameters $\{\theta_{i}, \psi_{i}, \phi_{aq}, \phi_{ev}\}$}
        \For{episode in episodes}{
            \For{timestep $t=1$ {\bfseries to} $T$}{
                $a_t \sim \pi_{\phi}\mleft(\cdot | s_t\mright)$, $s_{t} \sim p_{\theta_{i}}\mleft(\cdot | s_{t-1}, a_{t-1}, o_t\mright)$ \; 
                Observe environment transition and add to $\mathcal{D}$ \;
                }
            \For{trainstep $s=1$ {\bfseries to} $S$}{
                \For{particle $i=1$ {\bfseries to} $M$}{
                    Sample sequence batch $\{\mleft(o_t, a_t, r_t\mright)\}_{t=b}^{b+L} \sim D$\;
                    Compute states $s_{t} \sim p_{\theta_{i}}\mleft(s_t | s_{t-1}, a_{t-1}, o_t\mright)$\;
                    Estimate values $V_{\lambda, i}\mleft(s_{\tau}\mright) \leftarrow$ {\bfseries rollout}$(s_t, i)$\;
                    Representation learning of $\theta_i$ on $\{\mleft(r_{t}, o_{t+1}\mright)\}_{t=b}^{b+L}$\;
                    Regression of $\psi_i$ on $V_{\lambda, i}\mleft(s_{\tau}\mright)$ targets in~(\ref{eq:dreamer_objectives})
                }
                Sample sequences $\{\mleft(o_t, a_t, r_t\mright)\}_{t=b}^{b+L} \sim D$ \;
                \For{particle $i=1$ {\bfseries to} $M$}{
                    Compute states $s_{t} \sim p_{\theta_{i}}\mleft(s_t | s_{t-1}, a_{t-1}, o_t\mright)$\;
                    Estimate values $V_{\lambda, i}\mleft(s_{\tau}\mright) \leftarrow$ {\bfseries rollout}$(s_t, i)$\;
                }
                Compute ensemble statistics $\mu_{G}, \sigma_{G}$ via~(\ref{eq:rollout_moments})\;
                Update $\phi_{aq}$ by optimizing UCB objective in~(\ref{eq:ucb_objective_aquisition})\;
                Update $\phi_{ev}$ by optimizing mean returns in~(\ref{eq:mean_return})\;
            }
        }
    \caption{LOVE}
    \label{alg:love}%
  \end{algorithm}
\end{wrapfigure}
LOVE leverages an ensemble of latent variable models in combination with value function estimates to predict infinite-horizon trajectory performance and associated uncertainty. The acquisition policy is trained on a UCB objective to imagine positive futures and focus exploration on interactions with high predicted potential for long-term improvement. The ensemble members are constrained to operate over the same latent space to encourage learning of abstract representations conducive to the objective, while ensuring prediction consistency.%

\paragraph{Implementation}
The algorithm proceeds in two alternating phases.
In the online phase, the agent leverages its acquisition policy to explore interactions optimistically and resulting transitions are appended to memory $\mathcal{D}$. 
In the offline phase, the agent first updates its belief about environment behavior and then adjusts its policy accordingly.
The representation learning step extends the procedure introduced in \citet{hafner2019dream} to ensemble learning and groups each model with a unique value function estimator into a particle. The batch order is varied between particles during training to ensure variability in the gradient updates and to prevent collapse into a single mode.
The policy learning step combines particle predictions to generate value targets in Eq.~(\ref{eq:particle_return}) by simulating ensemble rollouts from the same initial conditions.
The trajectory return statistics of Eq.~(\ref{eq:rollout_moments}) and Eq.~(\ref{eq:mean_return}) are combined into the UCB objective of Eq.~(\ref{eq:ucb_objective_aquisition}), which the acquisition policy optimizes, while the evaluation policy optimizes the mean return in Eq.~(\ref{eq:mean_return}).
\section{Experiments}
We evaluate LOVE on continuous visual control tasks. 
First, we showcase intrinsic exploration in the absence of reward signals.
Then, we illustrate interpretable uncertainty estimation of long-term returns with sparse reward signals.
Finally, we benchmark performance on tasks from the DeepMind Control Suite \citep{deepmindcontrolsuite2018}. 
We use a single set of hyperparameters throughout, as detailed in Appendix~\ref{app:parameters}.%
\subsection{Intrinsic exploration without reward feedback}
\label{sec:bug_trap}

\begin{figure}[t!]
\centering
    \includegraphics[width=1.0\linewidth]{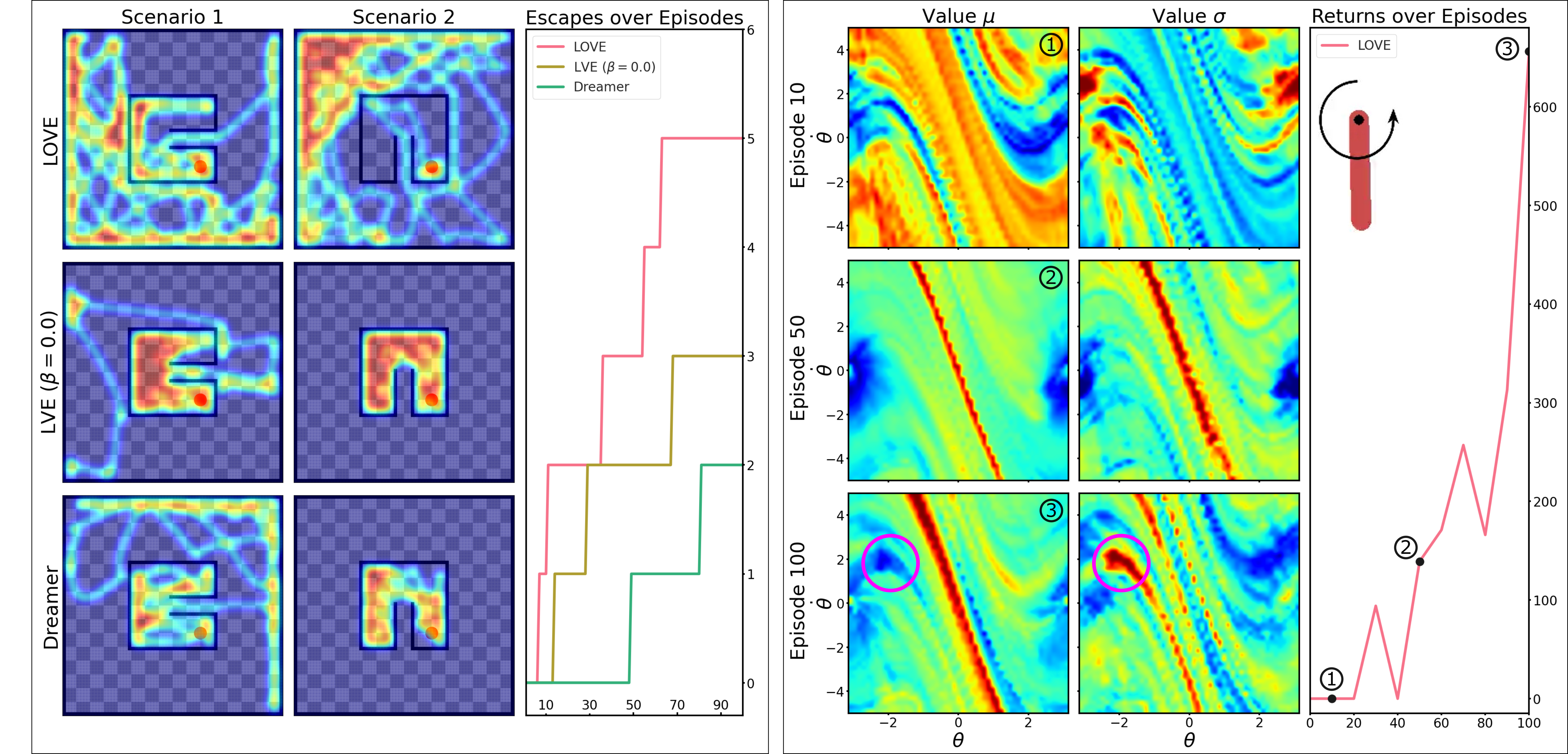}
\caption{\textbf{Left} - the agent starts inside a bug trap (red dot) and does not receive reward feedback. We provide occupancy maps for LOVE, LVE (no optimism), and Dreamer with escape counts on $3$ seeds. LOVE's ability to consider uncertain long-term performance enables consistent exploration in the absence of reward signals yielding the highest escape rate and area coverage. \textbf{Right} - Pendulum initialized at the bottom with sparse rewards at the top. Predicted ensemble mean (left) and standard deviation (right) of the value function. The agent first learns to quantify model uncertainty (Ep.~10), and then discovers high reward potential at the top (Ep.~50). Once the agent learned to reach the top, remaining uncertainty with low potential for improvement is not further explored (Ep.~100, circle).}
\label{fig:pointmass_pendulum}
\end{figure}%

In the absence of reward feedback an agent relies on intrinsic motivation to explore.
LOVE enables intrinsic exploration by considering uncertainty over infinite-horizon returns.
Without informative mean performance estimates, optimism acts as a proxy to guide exploration (see Eq.~\ref{eq:ucb_objective_aquisition}).
We consider a reward-free bug trap environment. The agent starts inside the trap and can exit through a narrow passage. There is no reward feedback to guide planning and escapes need to arise implicitly through exploration. 
The agent is represented as a pointmass under continuous actuation-limited acceleration control and observes a top-down view of the environment. 
The ratio of agent diameter to passage width is $0.9$ and collisions are modelled as inelastic with the coefficient of restitution set to $0.3$. 
The relative size constrains the set of policies allowing for entry into the passage, while inelastic collisions reduce random motions.%

We compare performance of LOVE to both Dreamer and LVE, a variation that ablates on LOVE's optimism ($\beta=0$).
Two variations of the environment are run on $3$ random seeds for $100$k timesteps.
Exemplary occupancy maps are provided in Figure~\ref{fig:pointmass_pendulum} (left). We note that LOVE escapes the bug trap in both scenarios to explore a much larger fraction of the state space than either LVE or Dreamer.
This can be attributed to LOVE's ability to envision uncertain rewards beyond the preview horizon, which optimistically drives deep exploration in the absence of reward signals (row 1). 
Removing long-term optimism by only guiding interactions through mean performance estimates can lead to prediction collapse and the agent assuming the absence of reward without explicitly querying the environment for confirmation (row 2). 
We observe similar behavior for the Dreamer agent, which employs a single latent model and leverages random noise for exploration (row 3).
The total escape count confirms this insight (column 3), where the remaining occupancy maps are provided in Appendix~\ref{app:bugtrap_extended}.
\begin{wrapfigure}[11]{l}{0.40\linewidth}
    \includegraphics[width=1.0\linewidth]{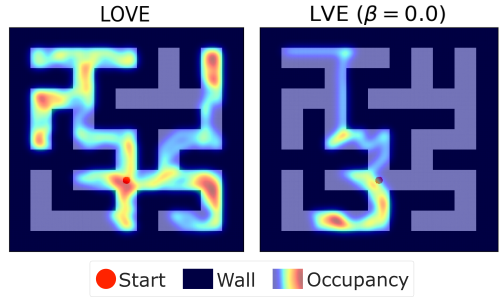}
  \caption{Reward-free maze exploration.}
  \label{fig:maze_main}
\end{wrapfigure}
We further highlight the differences in intrinsic exploration between LOVE and LVE on a reward-free maze exploration task. The environment follows the same physics as the bug trap and the ratio of agent diameter to corridor width is $0.33$. The occupancy maps in Figure~\ref{fig:maze_main} exemplify how optimism over returns can set its own exploration goals in domains without reward feedback. 
These observations further underline the potential of combining model ensembling for policy debiasing with optimism over infinite-horizon returns in guiding deep exploration.

\subsection{Targeted exploration with sparse rewards}
Efficient planning under sparse reward feedback requires long-term coordination.
LOVE focuses on interactions with high potential for improving infinite-horizon performance by trading-off predicted returns with their uncertainty.
The agent then only reduces uncertainty when it aligns with successful task completion.
We consider a sparse reward pendulum task, as its functional representations are readily interpretable.
The pendulum is always initialized in a downward configuration and only receives reward at the top.
The agent can therefore not rely on random initialization to observe initial reward signals and needs to actively explore the upright configuration.
Figure~\ref{fig:pointmass_pendulum} (right) visualizes the mean and variance of the value ensemble during training.
We generate predictions in state-space by providing $5$ contextual images for velocity estimation. 
Initially, the agent builds a world model within which it learns to quantify uncertainty (Ep.~10). 
The agent then leverages the model to explore the top. This yields a spinning behavior and reward propagates into the value mean, while uncertainty is allocated for trajectories that could yield stabilization at the top (Ep.~50).
The agent queries these promising motions and refines its behavior to reach the upright, reducing associated uncertainty (Ep.~100).
The UCB objective allows the agent to ignore remaining uncertainty tangential to successful task completion (circle).
LOVE therefore guides exploration via potential for long-term improvement while offering interpretable uncertainty estimates.

\subsection{Performance with mixed sparse and dense rewards}
The previous sections highlighted LOVE's intrinsic motivation to explore potential for long-term improvement based on its uncertainty estimation over returns. %
In the following, we compare performance on a variety of visual control tasks from the DeepMind Control Suite and sparsified variations.
The environments feature $\mleft(64, 64\mright)$ image observations, continuous actions, and dense or sparse rewards.
The sparse Cheetah Run, Walker Run and Walker Walk tasks are generated by zeroing rewards below a threshold ($0.25$, $0.25$ and $0.7$, respectively; rescaled to [0, 1]).
We use the same set of parameters throughout all experiments mirroring those of Dreamer. However, we do not use random exploration noise.
We furthermore use an ensemble of $M=5$ latent models, an initial UCB trade-off of $\beta=0.0$, an episodic growth of $\delta=10^{-3}$, and alter the policy learning rate and training steps to account for the more complex UCB objective. %
The changes to the policy learning rate and training steps are also applied to Dreamer to yield $\Delta$Dreamer, as we found this to improve performance (see Appendix~\ref{app:ablation_dreamer}).
We furthermore introduce an exploration baseline, $\Delta$Dreamer$+$Curious, inspired by the curiosity bonus in \citet{sekar2020planning} over an ensemble of RSSM dynamics.
Table~\ref{tbl:dmc_main} provides benchmarking performance on $5$ seeds over $300$ episodes (see also Appendix~\ref{sec:app_dmc_benchmark}).
On the sparsified environments, LOVE achieves $40\%$ average improvement over the state-of-the-art $\Delta$Dreamer agent and $30\%$ over both the exploration baseline $\Delta$Dreamer$+$Curious and ensembling baseline LVE. On the other environments, LOVE improves by $26\%$, $37\%$, and $9\%$, respectively.
This suggests that LOVE's combination of latent model ensembling with directed exploration can aid in identifying interactions conducive to task completion.
Ensembling reduces propagation of model-specific biases into the policy.
Optimistic exploration of uncertain long-term returns focuses sampling on regions with promising performance estimates while ignoring uncertain regions that are tangential to task completion.
This is an advantage of formulating the exploration objective in terms of uncertain rewards and not uncertain dynamics, as shown by the comparison to $\Delta$Dreamer$+$Curious.
We ablate the performance of LOVE on the UCB trade-off parameter $\beta$ and compare against LVE ($\beta=0$).
We note that while LOVE outperforms LVE on the majority of tasks, both reach similar terminal performance in several instances. 
However, LOVE provides clear improvements on the fully-sparse Cartpole Swingup and the partially-sparse Hopper Hop tasks, as well as the sparse locomotion tasks (Table~\ref{tbl:dmc_main}, top). These environments initialize the agent in configurations that provide no reward feedback, thereby forcing the agent to actively explore. LOVE leverages uncertainty-guided exploration and gains an advantage under these conditions.
Similarly, this can explain performance on the Pendulum Swingup task. While the task only provides sparse reward feedback, random environment initializations offer sufficient visitation to the non-zero reward states, removing the need for active exploration. 
LOVE also improves performance on the dense Walker tasks. These environments require learning of stable locomotion patterns under complex dynamics, where directed exploration efficiently identifies tweaks to the gait.
Similar to the bug trap environment of Section~\ref{sec:bug_trap}, we observe that optimistic exploration is especially favoured by objectives that provide sparse reward feedback, while enabling efficient discovery of tweaks to motion patterns under complex dynamics.
We additionally compare to Data Regularized Q (DrQ), a concurrent model-free approach. DrQ updates its actor-critic models online giving it an advantage over LOVE and Dreamer. 
LOVE performs favourably on the majority of environments with significant differences on the sparse tasks (Pendulum, Cartpole Sparse) and the locomotion tasks (Hopper, Walker). DrQ outperforms LOVE on Finger Spin and Cartpole with dense rewards. On these tasks, learning an explicit world model may actually be detrimental to attaining performance quickly. The former task features high-frequency behaviors that may induce aliasing, while the latter task allows the agent to leave the frame yielding transitions with no visual feedback. We additionally provide converged performance results for pixel-based D4PG \citep{barth2018distributed} and proprioception-based A3C \citep{mnih2016asynchronous} at $10^8$ environment steps to put the results into perspective.%

\begin{table*}[t]
    \small
    \centering
    \begin{tabular}{l c c c c c | c c}
    \hline
        300k steps & LOVE & LVE & $\Delta$Dreamer & $+$Curious & DrQ & A3C & D4PG\\
    \hline
        Cheetah Sparse (R) & $614 {\scriptstyle\pm 262}$ & $606 {\scriptstyle\pm 117}$ & $259 {\scriptstyle\pm 280}$ & $514 {\scriptstyle\pm 302}$ & $-$ & $-$ & $-$\\
        Walker Sparse (R) & $234 {\scriptstyle\pm 129}$ & $164 {\scriptstyle\pm 151}$ & $78 {\scriptstyle\pm 60}$ & $106 {\scriptstyle\pm 120}$ & $-$ & $-$ & $-$\\
        Walker Sparse (W) & $935 {\scriptstyle\pm 24}$ & $864 {\scriptstyle\pm 135}$ & $584 {\scriptstyle\pm 315}$ & $639 {\scriptstyle\pm 277}$ & $-$ & $-$ & $-$\\
        Hopper Sparse (S) & $831 {\scriptstyle\pm 72}$ & $778 {\scriptstyle\pm 142}$ & $649 {\scriptstyle\pm 212}$ & $530 {\scriptstyle\pm 246}$ & $-$ & $-$ & $-$\\
    \hline
        Avg. Diff. to LOVE & +0\% & -11\% & -46\% & -35\% & $-$ & $-$ & $-$\\
    \hline
    \hline
        Cartpole Dense & $688 {\scriptstyle\pm 124}$ & $666 {\scriptstyle\pm 78}$ & $512 {\scriptstyle\pm 253}$ & $539 {\scriptstyle\pm101}$  & $781 {\scriptstyle\pm 100}$ & $558 {\scriptstyle\pm 7}$ & $862 {\scriptstyle\pm1}$ \\
        Cartpole Sparse & $631 {\scriptstyle\pm 259}$ & $192 {\scriptstyle\pm 132}$ & $388 {\scriptstyle\pm 245}$ & $219 {\scriptstyle\pm 199}$ & $231 {\scriptstyle\pm 337}$ & $180 {\scriptstyle\pm 6}$ & $482 {\scriptstyle\pm 57}$\\
        Cheetah Run & $771 {\scriptstyle\pm 118}$ & $753 {\scriptstyle\pm 120}$ & $655 {\scriptstyle\pm 93}$ & $760 {\scriptstyle\pm 73}$ & $533 {\scriptstyle\pm 143}$ & $214 {\scriptstyle\pm 2}$ & $524 {\scriptstyle\pm 7}$\\
        Finger Spin & $584 {\scriptstyle\pm 302}$ & $605 {\scriptstyle\pm 306}$ & $348 {\scriptstyle\pm 156}$ & $380 {\scriptstyle\pm 173}$ & $898 {\scriptstyle\pm 131}$ & $129 {\scriptstyle\pm 2}$ & $986 {\scriptstyle\pm 1}$\\
        Hopper Hop & $203 {\scriptstyle\pm 104}$ & $165 {\scriptstyle\pm 96}$ & $118 {\scriptstyle\pm 95}$ & $83 {\scriptstyle\pm 87}$ & $151 {\scriptstyle\pm 51}$ & $1 {\scriptstyle\pm 0}$ & $242 {\scriptstyle\pm 2}$\\
        Pendulum & $641 {\scriptstyle\pm 298}$ & $639 {\scriptstyle\pm 332}$ & $320 {\scriptstyle\pm 353}$ & $177 {\scriptstyle\pm 208}$& $399 {\scriptstyle\pm 298}$ & $49 {\scriptstyle\pm 5}$ & $681 {\scriptstyle\pm 42}$\\
        Walker Run & $528 {\scriptstyle\pm 123}$ & $506 {\scriptstyle\pm 67}$ & $438 {\scriptstyle\pm 85}$ & $451 {\scriptstyle\pm 119}$& $338 {\scriptstyle\pm 82}$ & $192 {\scriptstyle\pm 2}$ & $567 {\scriptstyle\pm 19}$\\
        Walker Walk & $947 {\scriptstyle\pm 36}$ & $914 {\scriptstyle\pm 52}$ & $902 {\scriptstyle\pm 94}$ & $873 {\scriptstyle\pm 122}$& $815 {\scriptstyle\pm 184}$ & $311 {\scriptstyle\pm 2}$ & $968 {\scriptstyle\pm 2}$\\
    \hline
        Avg. Diff. to LOVE & +0\% & -12\% & -29\% & -35\% & -18\% & -70\% & +9\%\\
    \hline
    \hline
    \end{tabular}
    \caption{DeepMind Control Suite. Mean and standard deviation on $9$ seeds after $3\times 10^5$ timesteps. LOVE improves sample efficiency particularly under sparse reward feedback. Temporally-extended optimism helps LOVE in actively exploring uncertain returns, providing an advantage over LVE. Formulating intrinsic motivation in reward-space enables LOVE to identify uncertain interactions conducive to solving the task, providing an advantage over the curiosity baseline $\Delta$Dreamer$+$Curious. 
    *D4PG, A3C: converged results at $10^8$ timesteps for reference.
    }
\label{tbl:dmc_main}
\end{table*}

\section{Conclusion}
We propose Latent Optimistic Value Exploration (LOVE), a model-based reinforcement learning algorithm that leverages long-term optimism to guide exploration for continuous visual control. 
LOVE leverages finite-horizon rollouts of a latent model ensemble in combination with value function estimates to predict long-term performance of candidate action sequences.
The ensemble predictions are then combined into an upper confidence bound objective for policy learning. 
Training on this objective yields a policy that optimistically explores interactions that have the potential of improving task performance while ignoring uncertain interactions tangential to task completion.
These aspects are particularly important for robot control in complex environments, where the engineer aims to provide sparse high-level feedback without intensive reward engineering and the system should efficiently learn without getting distracted by irrelevant environment behavior.
To this end, we evaluate LOVE regarding its exploration capabilities and performance on a variety of tasks. 
In the absence of reward signals, LOVE demonstrates an intrinsic motivation to explore interactions based on their information density.
Empirical results on DeepMind Control Suite tasks showcase LOVE's competitive performance and ability to focus exploration on interactions conducive to task completion, particularly when reward signals are sparse.
Lifting intrinsic motivation into reward-space is then preferred over dynamics-space, as shown by comparison to a curiosity baseline.
Furthermore, LOVE demonstrates improved sample efficiency over the recent model-based Dreamer agent and competitiveness with the model-free DrQ agent. 
%



\clearpage
%
\subsubsection*{Acknowledgements}
This work  was  supported  in  part  by  the  Office  of Naval Research (ONR) Grant N00014-18-1-2830, Qualcomm and Toyota  Research Institute (TRI). This article solely reflects the opinions and conclusions of its authors and not TRI, Toyota, or any other entity. We thank them for their support. The authors further would like to thank Lucas Liebenwein for assistance with cluster deployment, and acknowledge the MIT SuperCloud and Lincoln Laboratory Supercomputing Center for providing HPC resources. We would also like to thank the reviewers and program chairs for their helpful feedback and suggestions for improvement.


\bibliography{bib/bibliography}  


\newpage
\appendix

\newpage

\setcounter{section}{0}

\section{Parameters and implementation}
\label{app:parameters}
We use as single set of parameters throughout all experimental evaluations. The general model architecture follows \citet{hafner2019dream}, where the variational autoencoder from \citet{ha2018world} is combined with the RSSM from \citet{hafner2018planet}. We extend their default parameters by the ensemble size $M=5$, the initial UCB trade-off parameter $\beta_{ini}=0.0$, and the per-episode linear UCB growth rate $\delta=0.001$. The learning rates for the model, the value function and the policy are $6\times10^{-4}$, $8\times10^{-5}$, $2\times10^{-4}$, respectively, and updates are computed with the Adam optimizer \citep{kingma2014adam}. Throughout all experiments, the online phase consists of $1000$ environment interactions with an action repeat of $2$, while the offline phase consists of $200$ learning updates.

An overview of the range of hyper-parameter values that were investigated is provided in Table~\ref{tbl:parameters}. Not all possible pairings were considered and suitable combinations were determined by inspection, while the best pairing was selected empirically. Our implementation builds on Dreamer (\url{https://github.com/danijar/dreamer}) and the remaining parameters are set to their default values. Experiments were conducted on $4$ CPU cores in combination with $1$ GPU (NVIDIA V100). We will make the underlying codebase publicly available.
\begin{table}[H]
  \small
  \centering
  \begin{tabular}{l @{\hspace{1.5\tabcolsep}} c}
    \toprule
    \cmidrule(r){1-2}
    Param.     & Values \\
    \midrule
    LR$_{\pi}$ & $\mleft[8\times10^{-5}, \; 2\times10^{-4}\mright]$  \\
    Steps     & $\mleft[100, \; 200\mright]$ \\
    $\beta_{ini}$     & $\mleft[-0.1, \; 0.0, \; 0.1, \; 0.3, \; 0.5\mright]$ \\
    $\delta$     & $\mleft[+\{-10^{-3}, 0.0, 10^{-3}, 2\cdot10^{-3}\}, \times\{1.01, 1.015\}\mright]$ \\
    \bottomrule
  \end{tabular}
\caption{Hyper-parameters considered during training.}
\label{tbl:parameters}
\end{table}
\section{Network Architectures}
\label{app:architectures}
\begin{table*}
\small
\centering
  \begin{tabular}{l l l l}
   \toprule
    Layer Type $\qquad$ & Input (dimensions) & Output (dimensions) & Additional Parameters \\
   \toprule
    \mcTwo{Transition model (\textit{imagine 1-step})} & & \\
   \midrule
    Dense    & $s_{\tau-1,s}$ (30), $a_{\tau-1}$ ($n_a$) & $\text{fc}_{t,i}^1$ (200)   &  a=ELU        \\[1ex]
    GRU    & $\text{fc}_{t,i}^1$ (200), $s_{\tau-1,d}$ (200) & $\text{rs}_{\tau}$ (200),  $s_{\tau,d}$ (200)   &  a=tanh        \\[1ex]
    Dense    & $\text{rs}_{\tau}$ (200) & $\text{fc}_{t,i}^2$ (200)   &  a=ELU        \\[1ex]
    Dense    & $\text{fc}_{t,i}^2$ (200) & $\mu_{\tau,s}^{prior}$ (30), $\sigma_{\tau,s}^{prior}$ (30)   &  a=None        \\
   \toprule
    \mcTwo{Transition model (\textit{observe 1-step})} & & \\
   \midrule
    Dense    & $s_{\tau,d}$ (200), $z_{\tau}$ (1024) & $\text{fc}_{t,o}^1$ (200)   &  a=ELU        \\[1ex]
    Dense    & $\text{fc}_{t,o}^1$ (200) & $\mu_{\tau,s}^{post}$ (30), $\sigma_{\tau,s}^{post}$ (30)   &  a=None        \\
   \toprule
    \mcTwo{Encoder model} & & \\
   \midrule
    Conv2D    & obs (64, 64, 3) & cv1 (31, 31, 32)   &  a=ReLU, s=2, k=(4,4)        \\[1ex]
    Conv2D & cv1 (31, 31, 32) &  cv2 (14, 14, 64)   &  a=ReLU, s=2, k=(4,4)        \\[1ex]
    Conv2D      & cv2 (14, 14, 64) &  cv3 (6, 6, 128)   &  a=ReLU, s=2, k=(4,4)  \\[1ex]
    Conv2D      & cv3 (6, 6, 128) &  cv4 (2, 2, 256) & a=ReLU, s=2, k=(4,4) \\[1ex]
    Reshape & cv4 (2, 2, 256) & $z_{\tau}$ (1, 1, 1024) & \\
   \toprule
    \mcTwo{Observation model} & & \\
   \midrule
    Dense     &  $s_{\tau,d}$ (200), $s_{\tau,s}$ (30) &  $\text{fc}_{o}^1$ (1, 1, 1024)   &  a=None      \\[1ex]
    Deconv2D    & $\text{fc}_{o}^1$ (1, 1, 1024) &  dc1 (5, 5, 128)   & a=ReLU, s=2, k=(5,5) \\[1ex]
    Deconv2D & dc1 (5, 5, 128) &  dc2 (13, 13, 64)   &  a=ReLU, s=2, k=(5,5)       \\[1ex]
    Deconv2D      & dc2 (13, 13, 64) &  dc3 (30, 30, 32) &  a=ReLU, s=2, k=(6,6)       \\[1ex]
    Deconv2D      & dc3 (30, 30, 32) & dc4 (64, 64, 3)   &  a=ReLU, s=2, k=(6,6)      \\
   \toprule
    \mcTwo{Reward model} & & \\
   \midrule
    Dense     & $s_{\tau,d}$ (200), $s_{\tau,s}$ (30) &  $\text{fc}_{r}^1$ (400)   &  a=ELU      \\[1ex]
    Dense $\times$ 1     & $\text{fc}_{r}^{\{1\}}$ (400) &  $\text{fc}_{r}^{\{2\}}$ (400)   &  a=ELU      \\[1ex]
    Dense     & $\text{fc}_{r}^2$ (400) &  $\text{fc}_{r}^3$ (1)   &  a=ELU   \\
   \toprule
    \mcTwo{Value model} & & \\
   \midrule
    Dense     &  $s_{\tau,d}$ (200), $s_{\tau,s}$ (30) &  $\text{fc}_{v}^1$ (400)   &  a=ELU      \\[1ex]
    Dense $\times$ 2     & $\text{fc}_{v}^{\{1, 2\}}$ (400) &  $\text{fc}_{v}^{\{2, 3\}}$ (400)   &  a=ELU      \\[1ex]
    Dense     & $\text{fc}_{v}^3$ (400) &  $\text{fc}_{v}^4$ (1)   &  a=ELU   \\
   \toprule
    \mcTwo{Action model} & & \\
   \midrule
    Dense     &  $s_{\tau,d}$ (200), $s_{\tau,s}$ (30) &  $\text{fc}_{a}^1$ (400)   &  a=ELU      \\[1ex]
    Dense $\times$ 3     & $\text{fc}_{a}^{\{1, 2, 3\}}$ (400) &  $\text{fc}_{a}^{\{2, 3, 4\}}$ (400)   &  a=ELU      \\[1ex]
    Dense     & $\text{fc}_{a}^4$ (400) &  $\mu_{a}$ ($n_a$), $\sigma_{a}$ ($n_a$) &  a=ELU   \\
   \bottomrule
  \end{tabular}
 \caption{General network architectures of the underlying models. We note that repeated layers have been condensed with Dense $\times$ $i$ referring to application of the same dense layer architecture $i$ times. Parameter abbreviations: a=activation, k=kernel, and s=stride. Adapted from \citet{hafner2019dream}.}
 \label{tbl:architecture}
\end{table*}
The base network architectures employed throughout this paper are provided in Table~\ref{tbl:architecture}. Each particle is assigned a distinct instance of its associated models. In the following, we briefly comment on how the two parts of the transition model interact and provide further insights into the remaining models.
\paragraph{Transition model}
The transition model follows the recurrent state space model (RSSM) architecture presented in~\citet{hafner2019dream, hafner2018planet}. The RSSM propagates model states consisting of a deterministic and a stochastic component, respectively denoted by $s_{t,d}$ and $s_{t,s}$ at time $t$. The stochastic component $s_{t,s}$ is represented as a diagonal Gaussian distribution. The transition model then leverages the \textit{imagine 1-step} method to predict priors for the associated mean and standard deviation, ($\mu_{t,s}^{prior}, \sigma_{t,s}^{prior}$), based on the previous model state and applied action. In the presence of observations, the \textit{observe 1-step} method can be leveraged to convert prior estimates into posterior estimates, ($\mu_{t,s}^{post}, \sigma_{t,s}^{post}$). The transition model may then propagate posteriors based on a context sequence using both the \textit{imagine 1-step} and \textit{observe 1-step} methods, from which interactions can be imagined by propagating prior estimates based on the \textit{imagine 1-step} method. Each particle uses a transition model that follows the presented network architecture, but possesses distinct parameters.

\paragraph{Encoder model} 
The encoder parameterization follows the architectural choices presented in~\citet{ha2018world}. The encoder generates embeddings based on 64$\times$64 RGB image observations.
\paragraph{Observation model} 
The observation model follows the decoder architecture presented in~\citet{ha2018world}. The image observations are reconstructed from the associated model states $s_{\tau}$.
\paragraph{Reward and value model}
Rewards and values are both predicted as scalar values from fully-connected networks that operate on the associated model states $s_{\tau}$, similar to~\citet{hafner2019dream}. Each particle uses a pairing of a reward model and a value model with distinct sets of parameters.
\paragraph{Action model}
The action model follows~\citet{hafner2019dream}, where the mean $\mu_{a}$ is rescaled and passed through a tanh to allow action saturation. It is combined with a softplus standard deviation based on $\sigma_{a}$ and the resulting Normal distribution is squashed via a tanh (see \citet{haarnoja2018sac}).

\section{Prediction uncertainty}
\label{app:uncertainty}
\begin{figure*}[t!]
\begin{center}
  \includegraphics[width=0.95\linewidth]{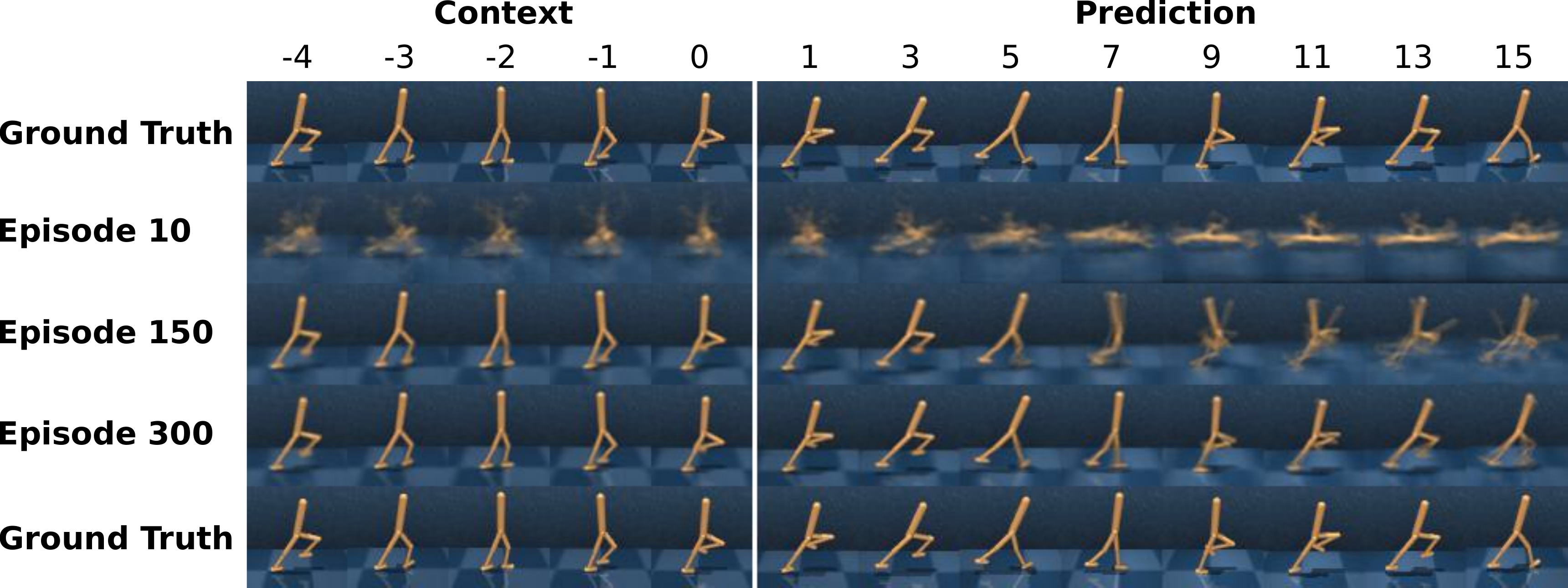}
\end{center}
\caption{Motion pattern of the Walker with low predictive uncertainty. The agent is provided with $5$ contextual images and predicts forward for $15$ steps (preview horizon), at different stages of training. The regular walking pattern is well-explored and only induces little deviation in the ensemble. This motion is desirable and the agent should focus on reducing its uncertainty over environment behavior.}
\label{fig:walker_certain}
\end{figure*}

\begin{figure*}[t!]
\begin{center}
  \includegraphics[width=0.95\linewidth]{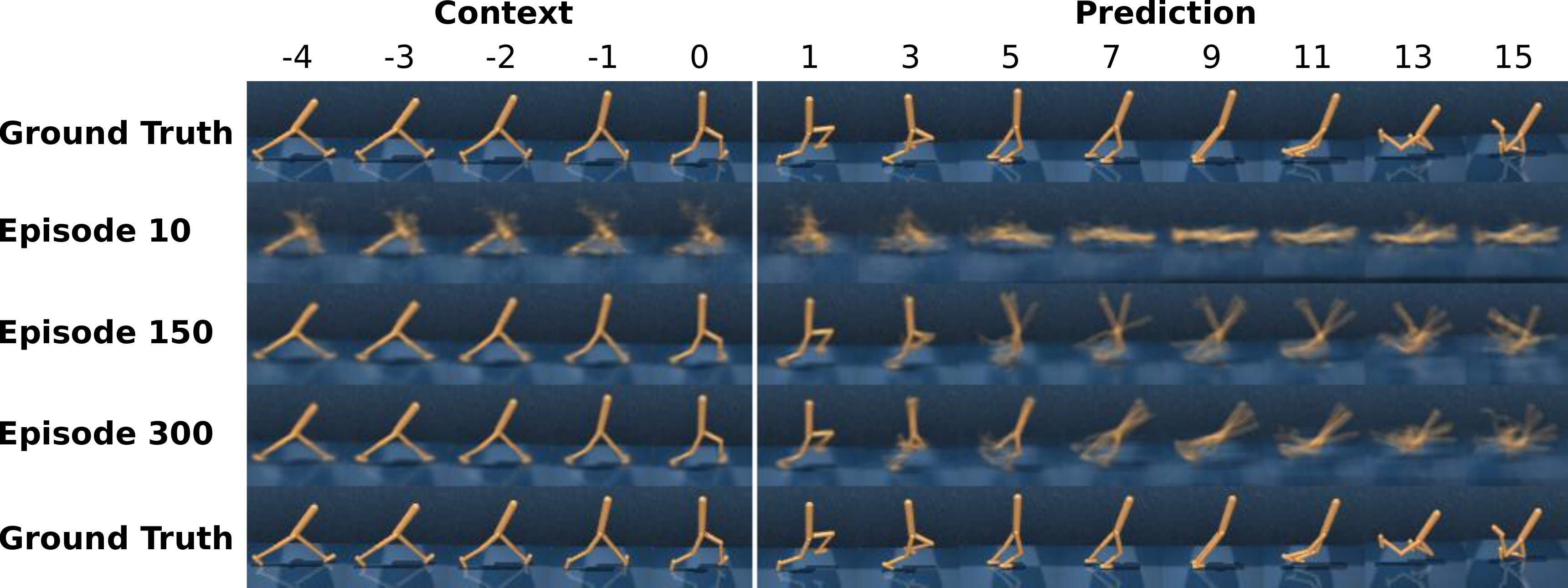}
\end{center}
\caption{Motion pattern of the Walker with high predictive uncertainty. The agent is provided with $5$ contextual images and predicts forward for $15$ steps (preview horizon), at different stages of training. The irregular falling pattern has not been extensively explored and high uncertainty remains in the ensemble. This motion is undesirable and the agent should not focus on reducing its uncertainty.}
\label{fig:walker_uncertain}
\end{figure*}

\begin{figure*}[t!]
\begin{center}
  \includegraphics[width=0.95\linewidth]{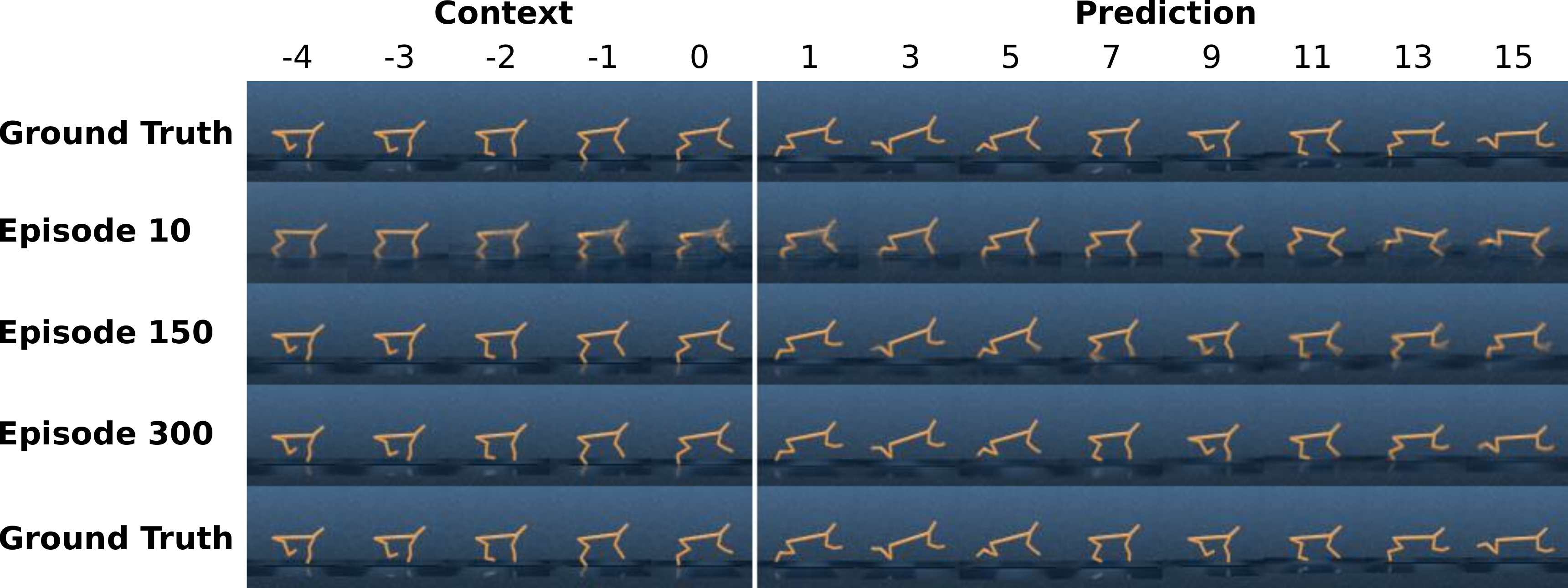}
\end{center}
\caption{Motion pattern of the Cheetah with low predictive uncertainty. The agent is provided with $5$ contextual images and predicts forward for $15$ steps (preview horizon), at different stages of training. The regular running pattern is well-explored and only induces little deviation in the ensemble. This motion is desirable and the agent should focus on reducing its uncertainty over environment behavior.}
\label{fig:cheetah_certain}
\end{figure*}

\begin{figure*}[t!]
\begin{center}
  \includegraphics[width=0.95\linewidth]{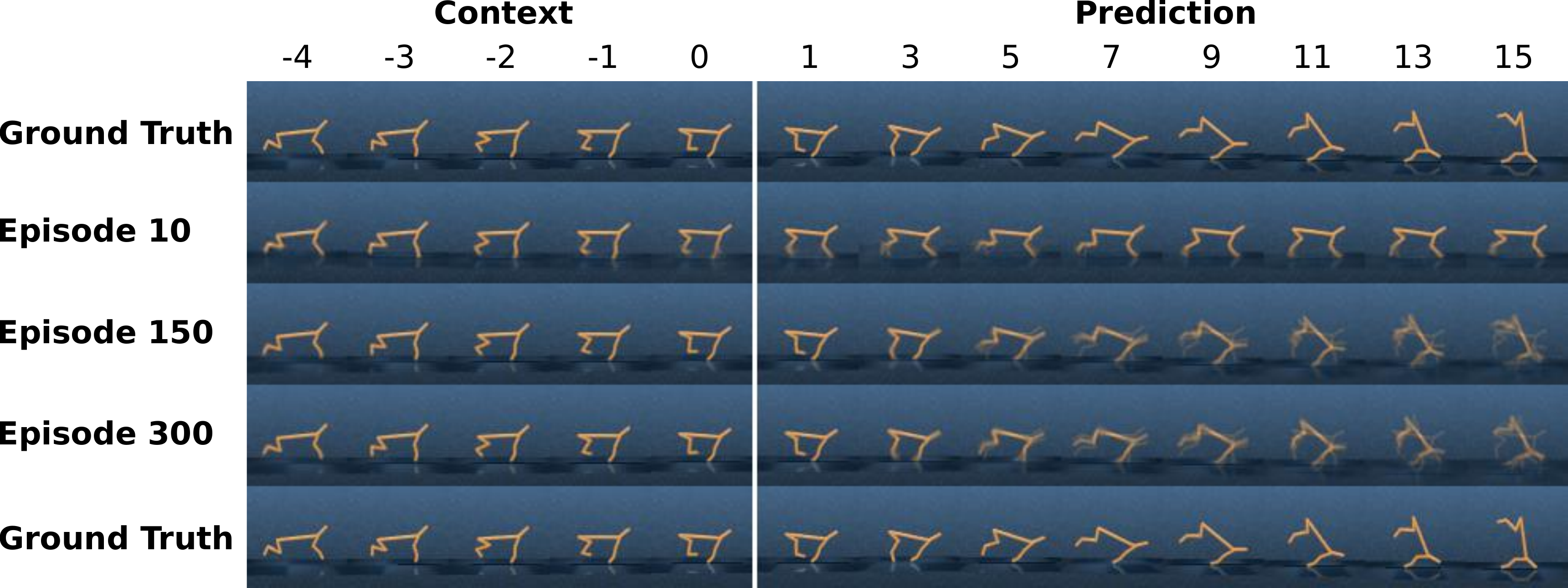}
\end{center}
\caption{Motion pattern of the Cheetah with high predictive uncertainty. The agent is provided with $5$ contextual images and predicts forward for $15$ steps (preview horizon), at different stages of training. The irregular falling pattern has not been extensively explored and uncertainty remains in the ensemble. This motion is undesirable and the agent should not focus on reducing its uncertainty.}
\label{fig:cheetah_uncertain}
\end{figure*}

We provide an illustrative visualization of how the prediction uncertainty in the ensemble evolves during model training. The ensemble is provided with context from a sequence of $5$ consecutive images and then predicts forward in an open loop fashion for $15$ steps (preview horizon). The ground truth sequence is compared to ensemble predictions after $10$, $150$, and $300$ episodes of agent training.%
Figures~\ref{fig:walker_certain} and \ref{fig:walker_uncertain} show two different motion patterns for the Walker Walk task. The motion in Figure~\ref{fig:walker_certain} can be described as a regular walking pattern. At the beginning of model training, the agent will have mostly observed itself falling to the ground and, in combination with a poorly trained policy, the ensemble predictions place the agent on the ground in a variety of configurations. After $150$ episodes, short-term uncertainty has been significantly reduced, while considerable uncertainty remains at the end of the preview window. After $300$ episodes, the ensemble predictions align with the ground truth sequence. The agent therefore focused on reducing uncertainty over this desirable motion pattern. This can be contrasted with the results of Figure~\ref{fig:walker_uncertain}, where uncertainty over an irregular falling pattern remains even after $300$ episodes. The falling motion is undesirable, and while the ensemble predictions agree on a fall being imminent, no significant amount of effort was spent on identifying exactly how the agent would fall. We can observe similar results on the Cheetah Run task for a running motion pattern in Figure~\ref{fig:cheetah_certain} and a falling motion pattern in Figure~\ref{fig:cheetah_uncertain}. However, the lower complexity Cheetah dynamics seem to allow for more precise predictions than on the Walker task.
\section{Baselines}
\label{app:baselines}
The baseline performance data for DrQ was taken from \citet{kostrikov2020image}, the ones for D4PG and A3C from \citet{deepmindcontrolsuite2018}, while the data for Dreamer was generated by running the official TensorFlow $2$ implementation of \citet{hafner2019dream}. It should be noted that both DrQ and D4PG use $84\times84$ image observations, whereas LOVE and Dreamer use $64\times64$ image observations. Larger resolution provides more fine-grained information, which potentially translates to improved planning. Furthermore, DrQ continuously refines its policy online, while the other algorithms only do so offline.
\section{Bugtrap extended}
\label{app:bugtrap_extended}
\begin{figure*}
\begin{center}
  \includegraphics[width=0.8\linewidth]{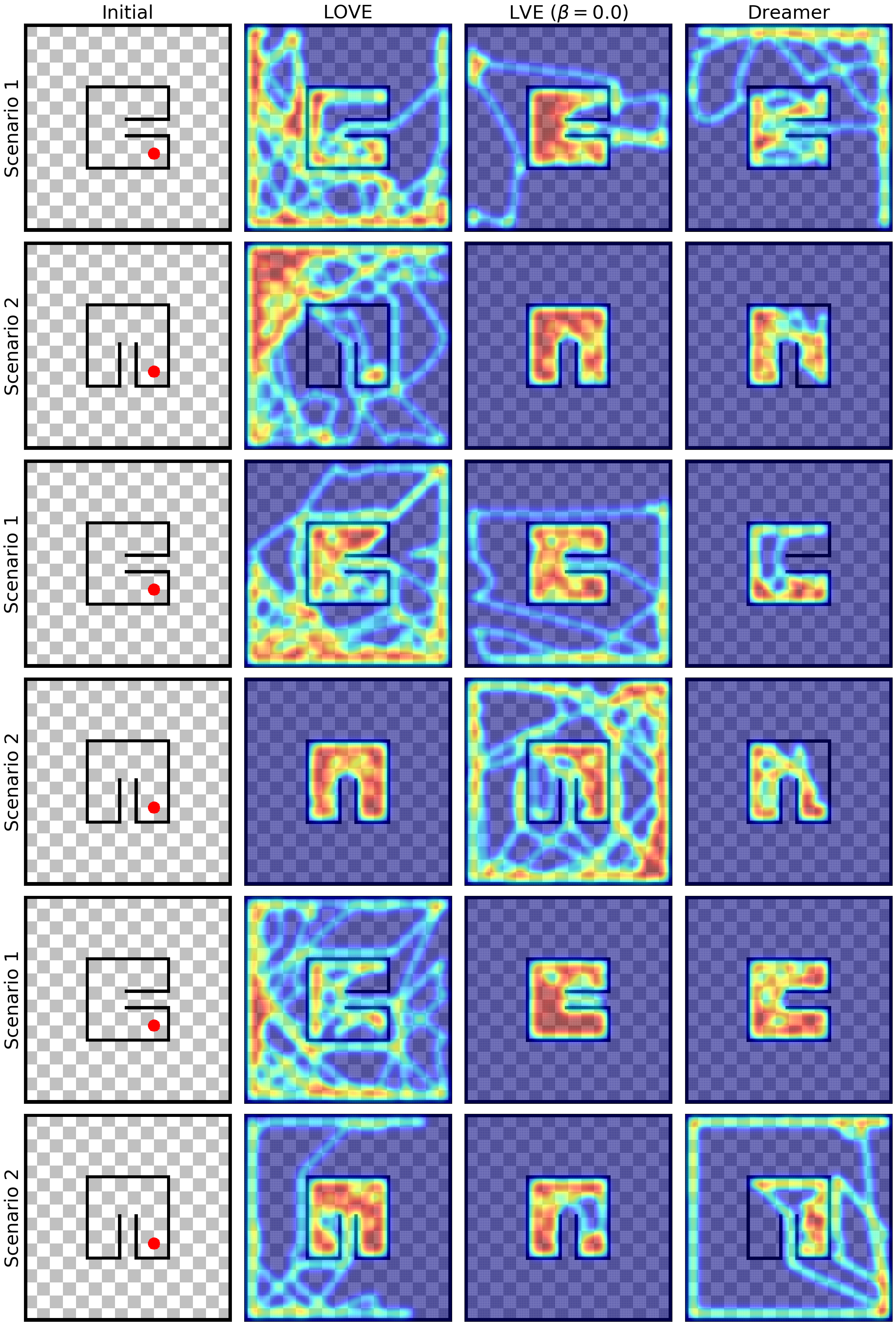}
\end{center}
\caption{Occupancy maps of the bug trap environment for two scenarios and three random seeds. In the absence of reward feedback, the uncertainty-guided exploration allows LOVE to escape during $5$ out of $6$ runs while achieving the highest area coverage in search of non-zero reward feedback. LVE removes optimistic exploration and as a result only escapes during $3$ runs, while significantly reducing area coverage. A similar pattern can be observed for the randomly exploring Dreamer agent.}
\label{fig:bugtrap_appendix}
\end{figure*}

We provide additional occupancy maps for the bug trap environment in Figure~\ref{fig:bugtrap_appendix}. The environment provides no reward feedback and assesses the agent's ability to actively search for informative feedback through intrinsic motivation. Furthermore, the environment geometry makes exploration of the outside area difficult. In the absence of useful mean performance estimates, LOVE leverages uncertainty-guided exploration to query interactions. This allows for escaping in $5$ out of $6$ trials and achieving the largest area coverage (column 2). LVE does not leverage uncertainty estimates and only escapes during $3$ trials (column 3), while displaying a highly reduced area coverage (rows 1 and 3). Similarly, random exploration allows the Dreamer agent to only escape in $2$ instances (column 4).

\section{Benchmarking on DeepMind Control Suite}
\label{sec:app_dmc_benchmark}
\begin{figure*}[t!]
\begin{center}
    \includegraphics[width=0.9\linewidth]{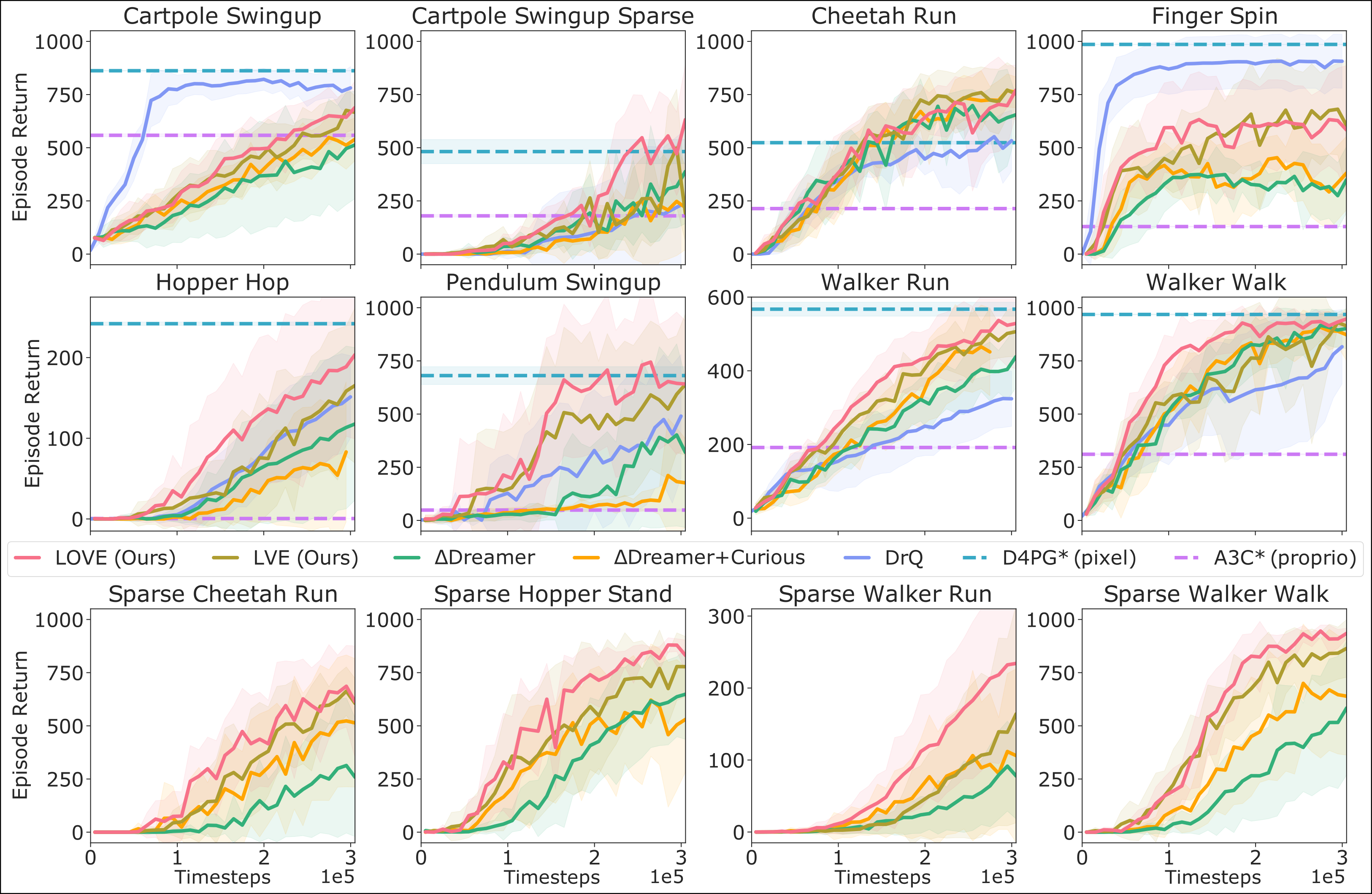}
\end{center}
\caption{DeepMind Control Suite. We evaluate performance over $300$ episodes on $9$ seeds. Solid lines indicate mean performance and shaded areas indicate one standard deviation. 
LOVE performs competitively and improves sample efficiency particularly under sparse reward feedback.
Temporally-extended optimism helps LOVE in actively exploring uncertain returns, providing an advantage over LVE.
Formulating intrinsic motivation in reward-space enables LOVE to identify uncertain interactions conducive to solving the task, providing an advantage over the curiosity baseline $\Delta$Dreamer$+$Curious. 
*D4PG, A3C: converged results at $10^8$ environment steps as reference.}
\label{fig:dmc_main}
\end{figure*}

Figure~\ref{fig:dmc_main} provides results for benchmarking performance over $300$ episodes. Performance is evaluated on $9$ seeds, where solid lines indicate the mean and shaded areas correspond to one standard deviation. LOVE's ability to explore uncertain long-term returns is particularly well-suited to reward structures that include sparsity. In particular, LOVE outperforms the curiosity baseline $\Delta$Dreamer$+$Curious as LOVE does not get distracted by uncertain but irrelevant unexpected environment behavior. Overall, LOVE yields the best performance across all tasks.
\section{Ablation study: Dreamer}
\label{app:ablation_dreamer}
\begin{figure*}
\begin{center}
  \includegraphics[width=0.95\linewidth]{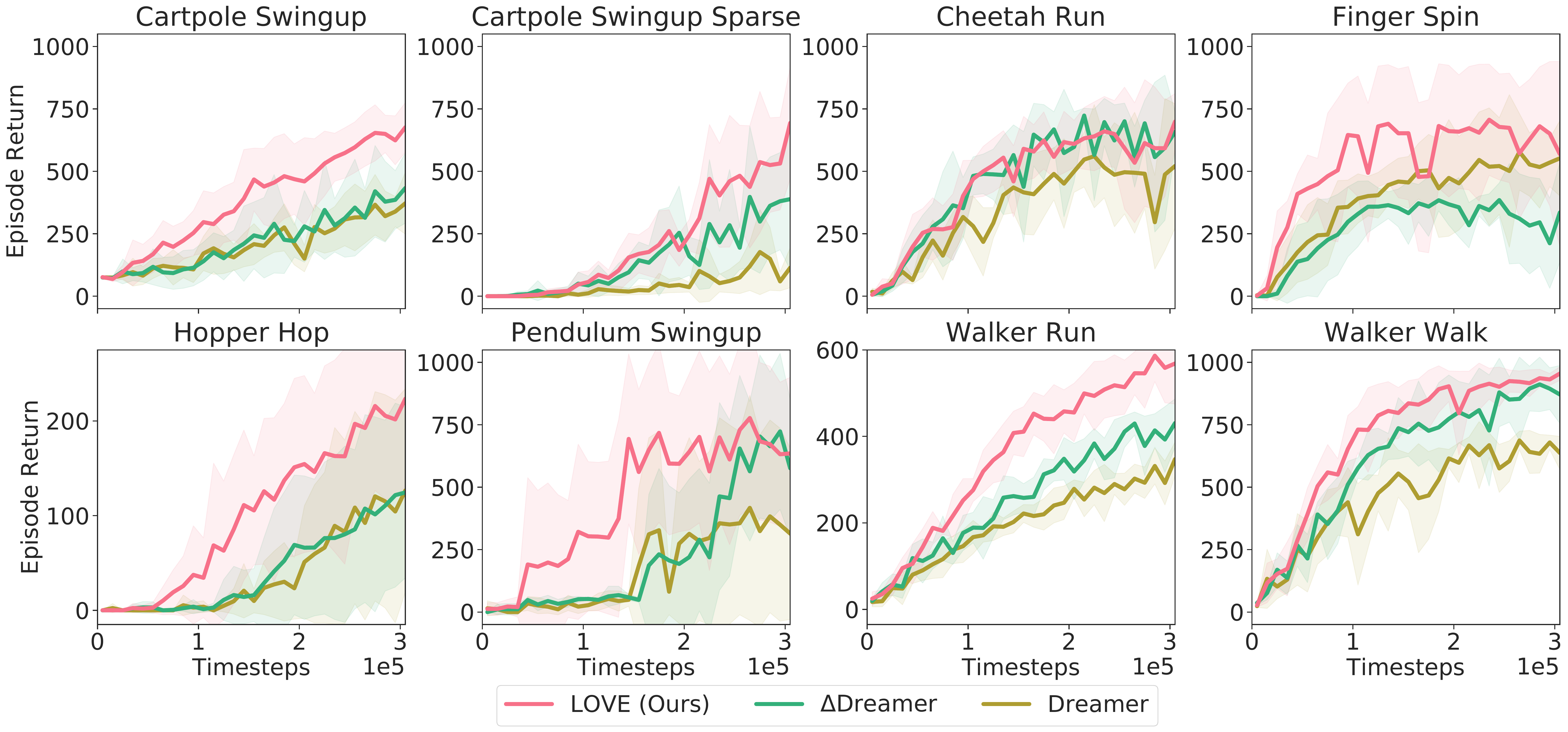}
\end{center}
\caption{Comparison to Dreamer with adapted policy learning rate and training steps ($\Delta$Dreamer). The changes improve performance of Dreamer on some environments, while significantly decreasing performance on the Finger Spin task. LOVE still outperforms $\Delta$Dreamer on the majority of tasks.}
\label{fig:dmc_ablation_dreamer}
\end{figure*}

We compare performance to $\Delta$Dreamer, a variation that uses our changes to the default parameters. Figure~\ref{fig:dmc_ablation_dreamer} indicates that performance improves on several tasks, while deteriorating on Finger Spin. LOVE outperforms $\Delta$Dreamer on the majority of tasks. It can thus be concluded that increased information propagation generally affects performance favourably. However, relying on a single model can propagate simulation bias into the policy and in turn impede efficient learning. This could serve as an explanation for the unchanged performance on the not fully observable Cartpole Swingup tasks, as well as the deteriorating performance on the high-frequency Finger Spin task.

\section{Ablation study: planning horizon}
\label{app:ablation_dreamer_planning_hor}
\begin{figure*}
\begin{center}
  \includegraphics[width=0.95\linewidth]{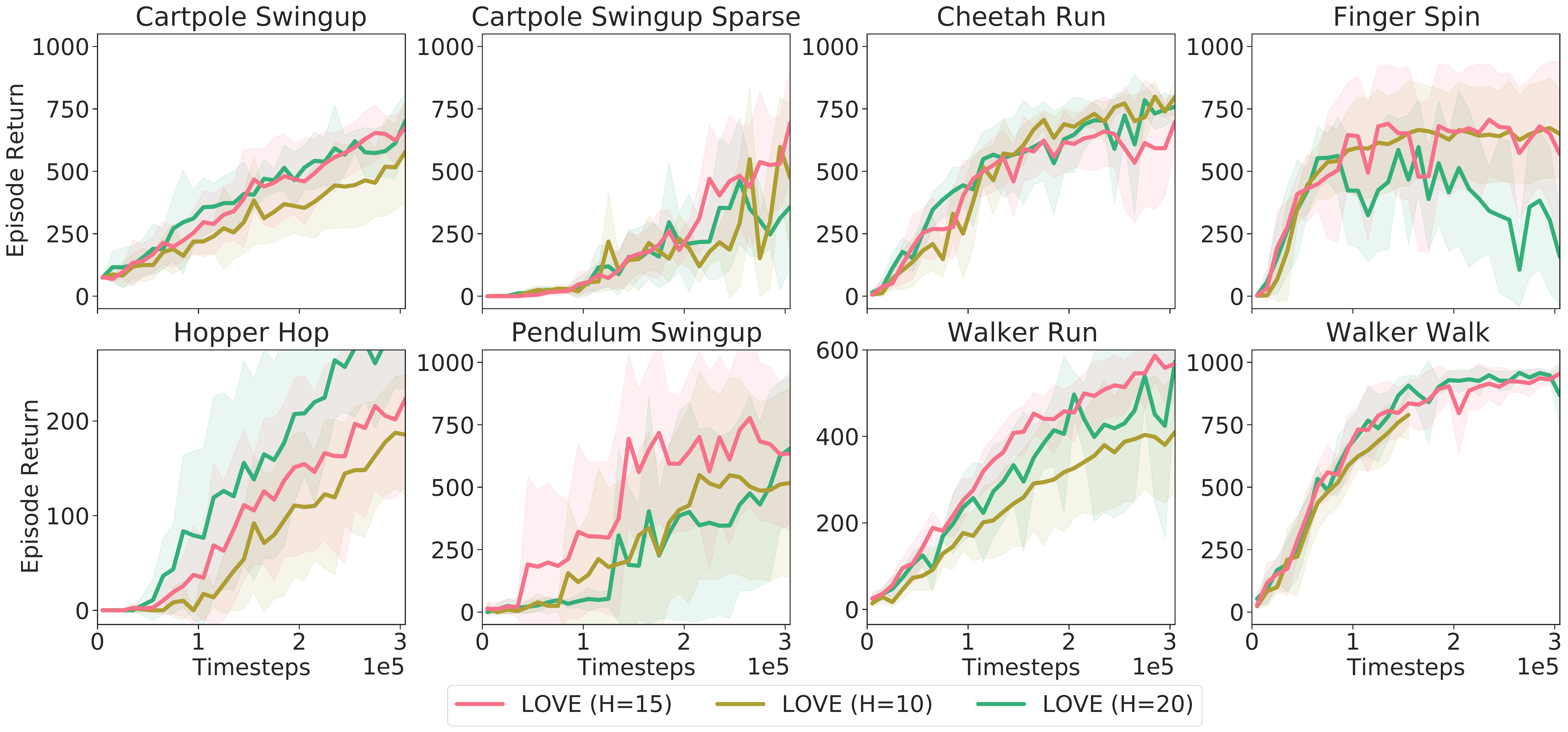}
\end{center}
\caption{LOVE under variation of the planning horizon.}
\label{fig:dmc_ablation_horizon_length}
\end{figure*}

We investigate increasing the planning horizon used during latent imagination. Longer horizons shift performance estimation from values towards rewards, which can be advantageous when the value function has not been sufficiently learned. Prediction quality over long horizons relies on accurate dynamics rollouts. Figure~\ref{fig:dmc_ablation_horizon_length} indicates that an intermediate horizon is a good trade-off. We note that the two Walker tasks had not completed at the time of submission, but provide a visible trend.
\section{Ablation study: $\beta$ schedule}
\label{app:ablation_dreamer_beta_schedule}
\begin{figure*}
\begin{center}
  \includegraphics[width=0.95\linewidth]{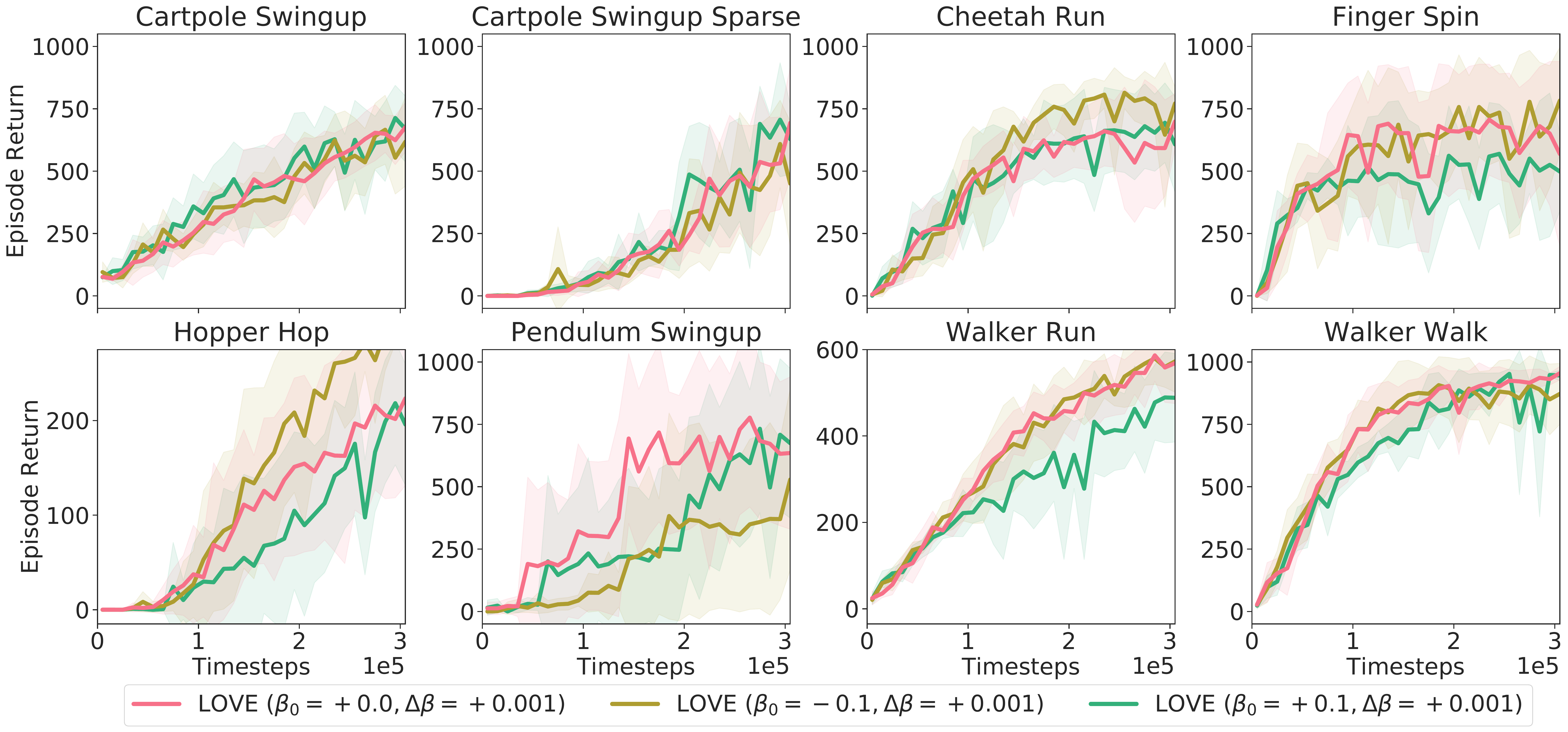}
\end{center}
\caption{LOVE under variation of the beta schedule.}
\label{fig:dmc_ablation_beta_schedule}
\end{figure*}

We investigate variations of the beta schedule, initializing either with a negative (initially pessimistic) or a positive value (initially optimistic). The former variation penalizes uncertainty in the beginning and then transitions to become optimistic, while the latter seeks out uncertainty from the start. Based on Figure~\ref{fig:dmc_ablation_beta_schedule}, we notice that terminal performance is mostly similar. The initially pessimistic agent exhibits reduced performance on the sparse pendulum, where it only explores well after it transitions to optimism (Episode 100), and improved performance on the challenging Hopper task, where initial pessimism potentially guards against local optima. Our choice of parameters tries to mitigate unfounded optimism during initialization (initial value 0), while encouraging exploration throughout the course of training (linear increase). 
Particularly at initialization, a strong positive UCB parameter may amplify random parameter noise as the agent will not have recovered meaningful representations or sufficiently propagated learned values, yet. 
Here, we chose the same values for all tasks, but one could imagine task-specific choices (negative beta for safe-RL, positive beta for optimistic exploration).

\section{Ablation study: ensemble size}
\label{app:ablation_dreamer_ens_size}
We investigate variation of the ensemble size. A smaller ensemble generates uncertainty estimates that are more susceptible to bias in the ensemble members and may even generate misleading estimates. Figure~\ref{fig:dmc_ablation_ensemble_size} demonstrates that a smaller ensemble (M=2) impacts performance unfavorably. We also provide data for a larger ensemble (M=10) on Cartpole Sparse, Cheetah, Pendulum and Walker Walk due to computation constraints. Generally, increasing the number of ensemble members increases the computational burden and we find the common literature choice of M=5 to perform sufficiently well.
\begin{figure*}
\begin{center}
    \includegraphics[width=0.95\linewidth]{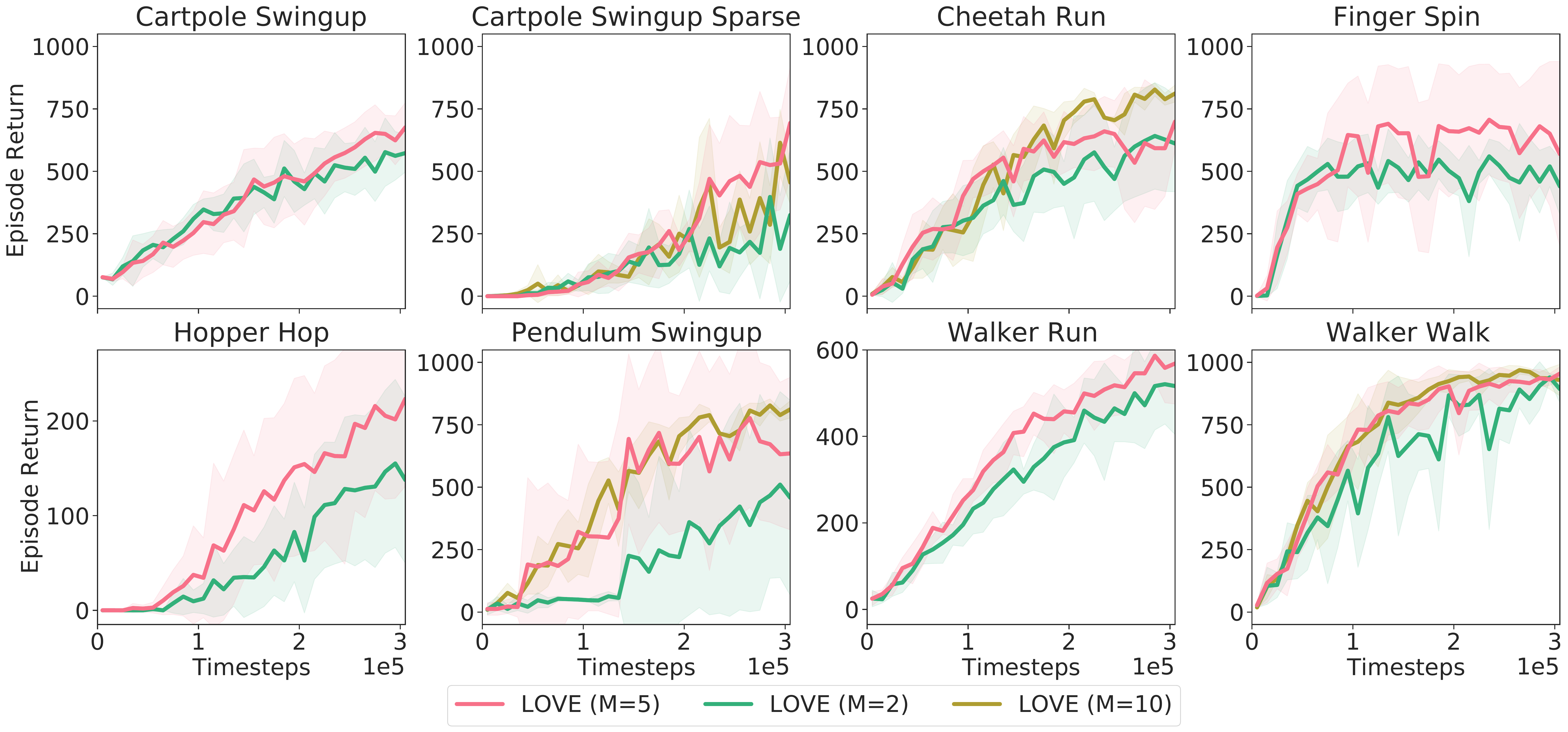}
\end{center}
\caption{LOVE under variation of the ensemble size.}
\label{fig:dmc_ablation_ensemble_size}
\end{figure*}%

\section{$\lambda$-returns}
\label{sec:lambda_return}
The value functions are trained with $\lambda$-return targets, which are computed according to
\begin{equation}
    \begin{aligned}
        \textstyle V_{\lambda}\mleft(s_{\tau}\mright) &\doteq \mleft(1 - \lambda\mright) \sum_{n=1}^{H-1} \lambda^{n-1} V_{N}^{n}\mleft(s_{\tau}\mright) + \lambda^{H-1} V_{N}^{H} \mleft(s_{\tau}\mright), \\
        \textstyle V_{N}^{k}\mleft(s_{\tau}\mright) &\doteq E_{q_{\theta}, q_{\phi}} \mleft( \sum_{n=\tau}^{h-1} \gamma^{n-\tau} r_{n} + \gamma^{h-\tau} v_{\psi} \mleft(s_{h}\mright) \mright),
    \end{aligned}
    \label{eq:value_estimates}
\end{equation}
with $h = \min \mleft(\tau+k, t+H\mright)$.

\section{Asymptotic performance}
We provide episode returns over $450$k environment steps to better assess asymptotic performance. We consider a subset of the environments across 4 seeds. We observe that LOVE converges to or beyond converged D4PG performance with the exception on Finger Spin. LOVE further significantly outperform the Dreamer baselines. Generally, the comparatively low performance of the model-based agents on Finger Spin could indicate that the required high frequency behavior can be difficult to learn based on an explicit model. Throughout, we employed the same UCB trade-off parameter schedule as described in Section~\ref{app:parameters} underlining that the agent does not get distracted by tangential uncertainty.
\begin{figure*}
\begin{center}
  \includegraphics[width=0.95\linewidth]{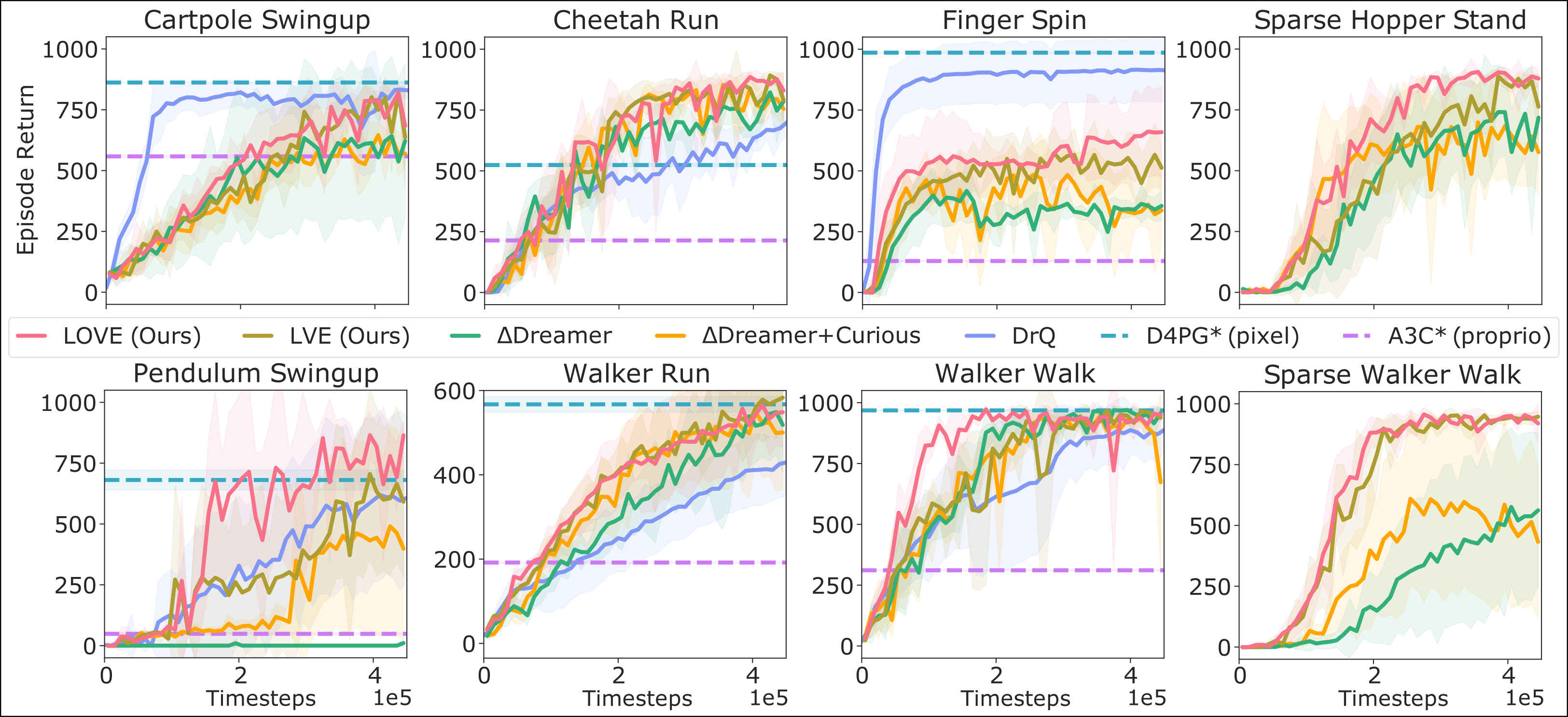}
\end{center}
\caption{Evaluation of asymptotic performance based $450$k environment steps. We observe that LOVE approaches or exceeds D4PG performance at $10^8$ environment steps on most environments.}
\label{fig:dmc_ablation_ensemble_size}
\end{figure*}%

\section{Interleaved Exploitation}
\begin{wrapfigure}{r}{0.5\textwidth}
\vspace{-20pt}
\begin{center}
    \includegraphics[width=1.00\linewidth]{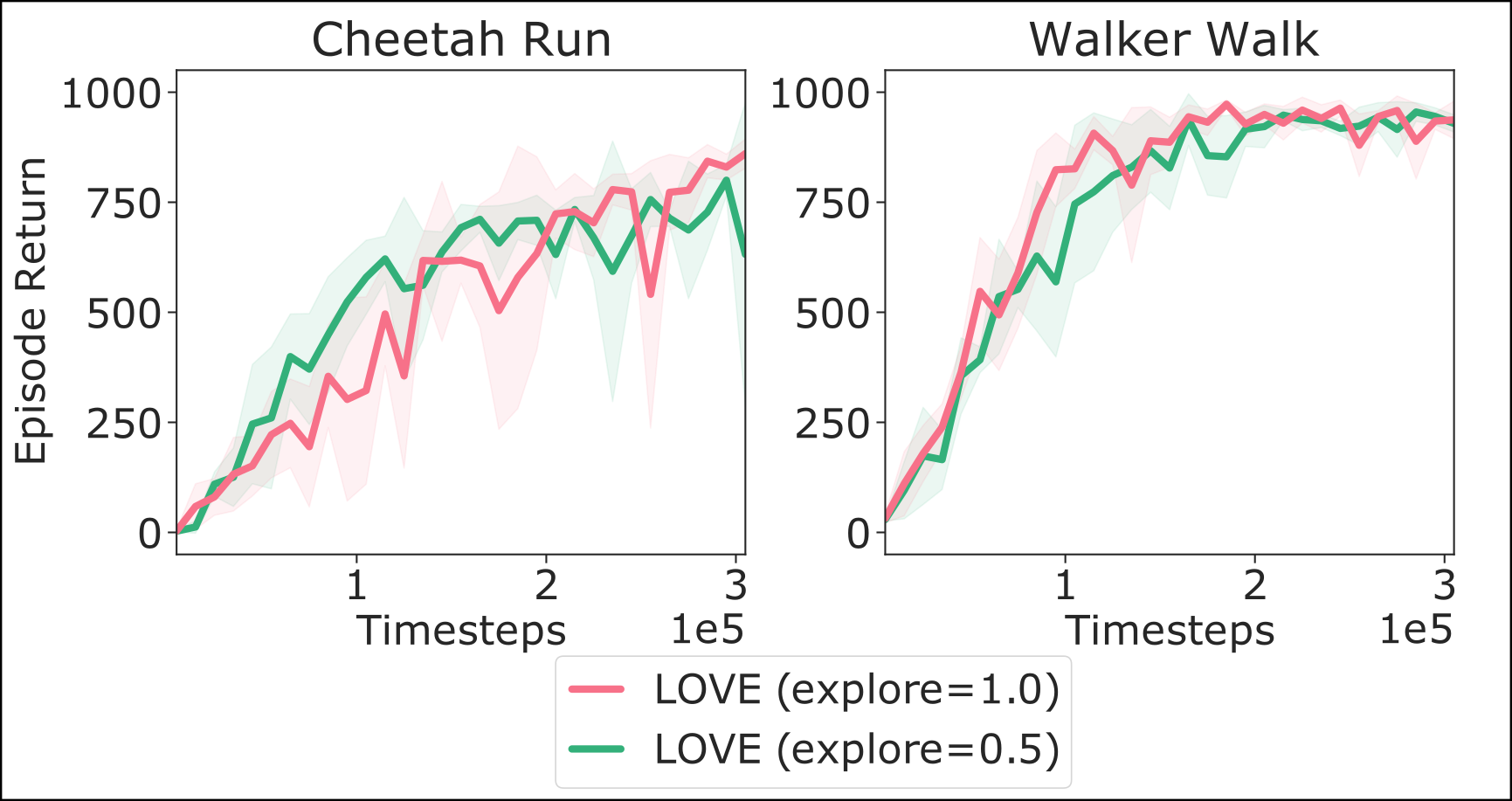}
\end{center}
\caption{
Interleaving action samples from the acquisition policy and the evaluation policy for generating environment interactions.
}
\label{fig:explicit_explore_exploit}
\vspace{-10pt}
\end{wrapfigure}%
We briefly evaluate explicitly interleaving the exploration and exploitation policy during learning, sampling from either policy with a probability of $p(\pi_{\phi_{\cdot}})=0.5$ with $\phi_{\cdot} \in \{\phi_{aq}, \phi_{ev}\}$. Due to computation constraints we limit ourselves to $4$ seeds on the Cheetah Run and Walker Walk tasks. From Figure~\ref{fig:explicit_explore_exploit}, we observe that LOVE's implicit exploration-exploitation trade-off (red) and interleaving explicit exploitation (green) perform similarly, with a slight edge for LOVE's implicit trade-off on the Walker task and for explicit exploitation on the Cheetah task. Generally, the effect of interleaving explicit exploitation to generate samples may depend on the nature of the task and the reward density.

\newpage
\section{Network Initialization}
\begin{wrapfigure}{r}{0.3\textwidth}
\vspace{-38pt}
\begin{center}
  \includegraphics[width=1.00\linewidth]{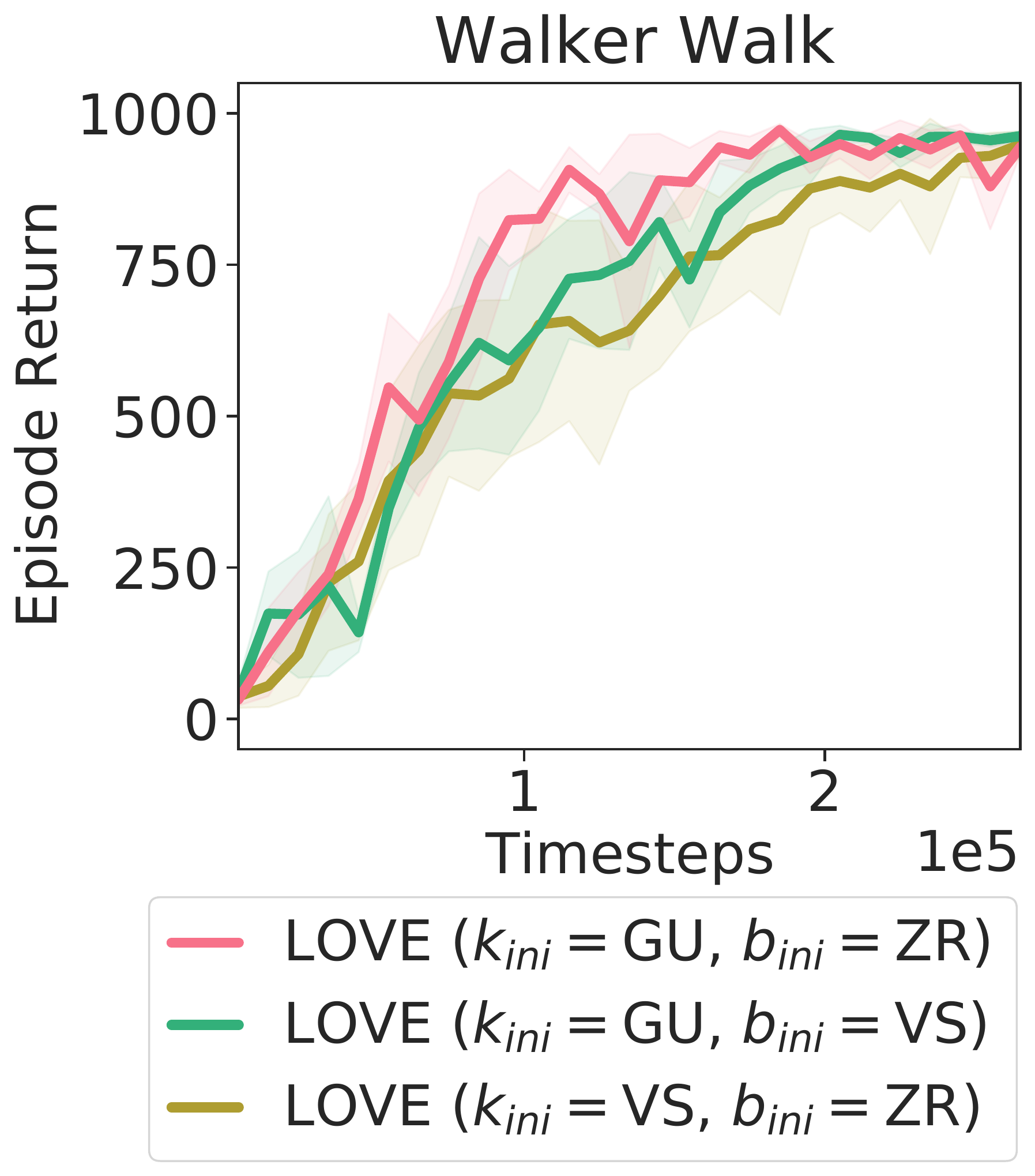}
\end{center}
\caption{
Variation of the kernel and bias initialization.
}
\label{fig:network_initialization}
\end{wrapfigure}%
We briefly evaluate changes to the kernel and bias initializers of the reward and value networks. The default initializers are Glorot-Uniform (GU) for the kernels and Zero (ZR) for the biases. For either, we consider the Variance-Scaling initializer (VS) as an alternative, where we use the scaling factor $\sigma=0.333$. We consider $4$ seeds each on the Walker Walk task. Based on Figure~\ref{fig:network_initialization}, we observe that changes to either the kernel or bias initializers slightly lower convergence speed while still reaching comparable asymptotic performance. More sophisticated initialization schemes may include explicit regularization of the network parameters towards random anchors as in~\cite{pearce2020uncertainty} and could harbor potential for further improving performance.

\section{Maze exploration}
\begin{wrapfigure}{r}{0.7\textwidth}
\vspace{-10pt}
\begin{center}
  \includegraphics[width=1.00\linewidth]{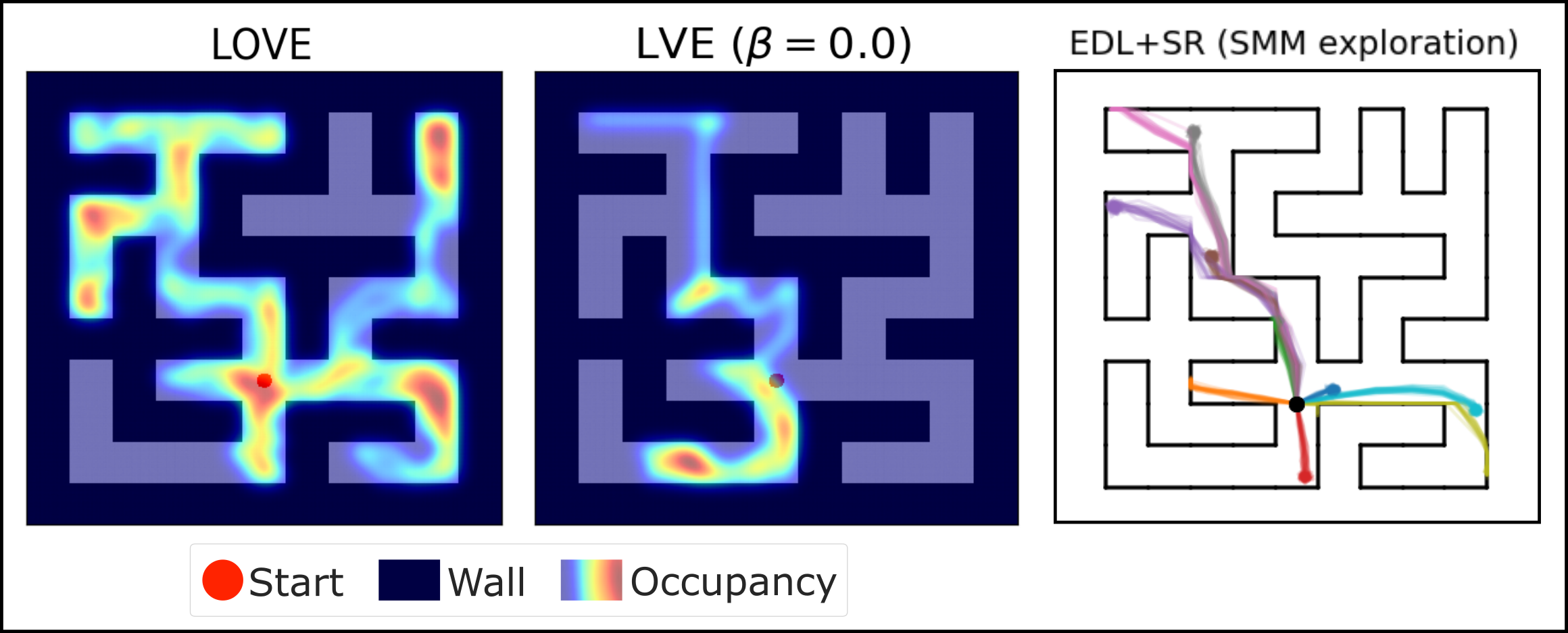}
\end{center}
\caption{
Maze exploration in comparison to the EDL~\cite{campos2020explore} agent.
}
\label{fig:maze_exploration}
\end{wrapfigure}%
We provide a qualitative comparison to the Explore, Discover and Learn (EDL) algorithm~\cite{campos2020explore} on the reward-free maze task. EDL differs from LOVE in its input-output modalities as EDL considers position control based on state input, whereas LOVE considers acceleration control based on image input. Modifying the EDL agent would be non-trivial and in order to provide fair qualitative insights we integrate our maze domain into their framework with position control from state observations. Particularly, we consider EDL with State Marginal Matching and Sibling Rivalry with the default hyperparameters. Based on Figure~\ref{fig:maze_exploration} (right), we see that the skills discovered by EDL mostly align with the occupancy trances of the LOVE agent, while both achieve greater coverage than LVE.


\end{document}